%% file: main.tex
\documentclass[10pt,twocolumn,letterpaper]{article}

\usepackage[pagenumbers]{cvpr} 

\usepackage{multirow, makecell}
\usepackage{graphicx}
\usepackage{float}
\usepackage[table]{xcolor}
\usepackage{amsmath, amssymb}
\usepackage{booktabs}
\usepackage{algorithm}
\usepackage{algpseudocode}

\usepackage{pifont}
\usepackage[dvipsnames]{xcolor}
\newcommand{\cmark}{\textcolor{ForestGreen}{\ding{51}}}
\newcommand{\xmark}{\textcolor{BrickRed}{\ding{55}}}

\input{preamble}
\definecolor{cvprblue}{rgb}{0.21,0.49,0.74}
\usepackage[pagebackref,breaklinks,colorlinks,allcolors=cvprblue]{hyperref}

\def\paperID{105} 
\def\confName{3DV\xspace}
\def\confYear{2027\xspace}

\title{SelectAnyTree: A Promptable Instance Segmentation Model \\ for 3D Forest LiDAR Point Clouds
\vspace{-10pt}
}

\author{
Trung Thanh Nguyen$^{1,2}$\thanks{Corresponding author: \texttt{nguyent@cs.is.i.nagoya-u.ac.jp}.}\hspace{5pt},
Daniel Lusk$^{3}$,
Kilian Gerberding$^{3}$,
Janusch Vajna-Jehle$^{3}$, \\
Tuan-Anh Vu$^{4}$, Duc Viet Le$^{5}$, Tu Vo$^{6}$, Phi Le Nguyen$^{7}$, \\ Yasutomo Kawanishi$^{8,2,1}$,
Takahiro Komamizu$^{1}$,
Ichiro Ide$^{1}$,
Julian Frey$^{3}$,
and Teja Kattenborn$^{3}$
\\
$^{1}$Nagoya University, Japan \hspace{5pt}
$^{2}$RIKEN, Japan  \\
$^{3}$University of Freiburg, Germany \hspace{5pt}
$^{4}$University of California, Los Angeles, USA \\
$^{5}$University of Twente, the Netherlands \hspace{5pt}
$^{6}$KC Machine Learning Lab, Korea \\
$^{7}$Hanoi University of Science and Technology, Vietnam  \hspace{5pt}
$^{8}$Ritsumeikan University, Japan
\vspace{-13pt}
}

\begin{document}
\maketitle
\input{sec/0_abstract}

\section{Introduction}
\label{sec:intro}
\input{sec/1_intro}

\section{Related work}
\label{sec:related_work}
\input{sec/2_related-work}

\section{SelectAnyTree}
\label{sec:proposed_method}
\input{sec/3_proposed-method}

\section{Experiments}
\label{sec:experimental_results}
\input{sec/4_evaluation}

\section{Conclusion}
\label{sec:conclusion}

\input{sec/5_conclusion}

\section*{Acknowledgment}
This study was supported by the Eva Mayr-Stihl Foundation and Deutsche Forschungsgemeinschaft (DFG) through the project LeafH2O (541018379) and Germany's Excellence Strategy (Future Forests–EXC-3127-533786343).

{
    \small
    \bibliographystyle{ieeenat_fullname}
    \bibliography{main}
}

\clearpage
\appendix
\input{sec/6_appendix}

\end{document}

%% file: sec/0_abstract.tex
\begin{abstract}
Instance segmentation of trees in forest LiDAR point clouds is constrained by label scarcity: A single hectare holds millions of points and hundreds of overlapping tree crowns, making manual annotation laborious, while automatic pre-segmentations offer no interactive refinement.
Inspired by the promptable paradigm of foundation segmentation models, we propose \textbf{SelectAnyTree}, which delineates any individual tree in a 3D forest point cloud from a few clicks and is purpose-built for promptable instance segmentation of 3D forest LiDAR scenes.
{The proposed SelectAnyTree couples three lightweight stages:
(1)~Sparse voxel scene encoder that embeds the forest \emph{once} into reusable features,
(2)~Click-to-query prompt encoder that turns each click into a single content query from its 3D position, positive/negative polarity, and the backbone feature of its nearest voxel, and
(3)~State-space query decoder that converts this query into one tree mask with linear-time complexity, with a mask feedback that conditions each refinement round on the previous mask.
Each additional tree therefore costs only a lightweight prompt-encoding and decoding pass, and the full model requires just 19.4\,M parameters, far fewer than prior promptable 3D models.
Additionally, we exploit forest-aware information by detecting treetops as local maxima of the Canopy Height Model (CHM) computed from the scene geometry, and associating one with the user's click as a free initial click.}
Across seven diverse forest regions and an independent held-out dataset, SelectAnyTree segments a target tree to 79.9 Intersection-over-Union (IoU) from a single click, 24.7 points above the strongest promptable baseline, and reaches every accuracy target with the fewest clicks.
The source code is available at \url{https://github.com/thanhhff/SelectAnyTree}.
\end{abstract}

%% file: sec/1_intro.tex
Forests are complex 3D ecosystems that play a central role in carbon storage, biodiversity, and climate regulation~\cite{white2016remote,kattenborn2021review}.
The combination of terrestrial or drone-based LiDAR with deep learning has made it possible to effectively translate the dense 3D structure of forests into quantitative information for inventory and ecological analysis~\cite{calders2020terrestrial,feigl2025close, kattenborn2021review}.
A core task underlying these applications is \emph{individual tree instance segmentation}; assigning every point of a forest scan to the tree to which it belongs.
Reliable per-tree delineation is a prerequisite for estimating biomass, crown architecture, and competition, as well as for downstream traits such as species or health status.

Recent forest instance segmentation methods have advanced rapidly, progressing from clustering-based pipelines~\cite{henrich2024treelearn,xiang2024automated} to end-to-end query-based architectures~\cite{kolodiazhnyi2024oneformer3d,xiang2025forestformer3d,nguyen_forestmamba}.
Despite their strong benchmark performance, these models share a common operating mode; they segment \emph{all} trees in a scene in a single fully-automatic forward pass.
This design has two practical consequences.
First, it provides no mechanism for a user to \emph{select} a particular tree of interest, whether to annotate a reference plot, extract a single specimen, or build training data, without running and post-processing the entire scene.
Second, in dense, multi-layered forest canopies where crowns are highly entangled and the models often fail~\cite{henrich2024treelearn,xiang2025forestformer3d}, there is no way to \emph{correct} an individual mistaken instance; errors can only be addressed by retraining or by costly manual editing.
In annotation-heavy ecological applications, where ground-truth masks are expensive and expert-driven, this lack of human control is a fundamental limitation.
At the same time, progress in forest intelligence from LiDAR remains limited by label scarcity, particularly for individual tree instances and downstream properties such as species or functional type~\cite{kattenborn2021review, xiang2025forestformer3d}.

In 2D vision, the well-known Segment Anything Model (SAM)~\cite{kirillov2023segment} reframed segmentation as a \emph{promptable} task, where a user specifies the object to be segmented through a few clicks and the model responds with a corresponding mask.
This paradigm of AI-assisted annotation has since been extended to 3D point clouds by interactive and promptable models such as AGILE3D~\cite{yue2024agile3d}, NPISeg3D~\cite{liu2025npising3d}, and Point-SAM~\cite{zhou2025pointsam}, and to native 3D part segmentation by PartSAM~\cite{zhu2026partsam}.
However, these models are designed for generic indoor objects, man-made 3D shapes, or object parts, and are not optimized to encode forest-specific structures.
Forest scenes differ sharply from these settings; they are large-scale and sparsely sampled, contain hundreds of trees with strongly overlapping but structurally diverse crowns, and exhibit a pronounced vertical ground-to-canopy organization that generic promptable models do not exploit.
As a result, directly applying object-centric promptable models to forest LiDAR leaves substantial domain structure unused.

To close this gap, we propose \textbf{SelectAnyTree}, a promptable model that interactively segments \emph{any} individual tree in real time from only a few user clicks.
{It couples three lightweight stages:
(1)~\emph{Sparse Voxel Scene Encoder} that embeds the forest \emph{once} into features reused by every prompt,
(2)~\emph{Click-to-Query Prompt Encoder} that fuses each click's 3D position, polarity, and the backbone feature of its nearest voxel into a single content query, disambiguating a lone click among densely entangled crowns, and
(3)~\emph{State-Space Query Decoder} that turns that query into one tree mask in linear time, with a mask feedback that conditions each refinement round on the previous mask.
Each additional tree therefore costs only a lightweight prompt-encoding and decoding pass.
Additionally, we exploit forest-aware information that generic promptable models ignore.
Local maxima of the Canopy Height Model (CHM), a raster storing the height of the tallest vegetation above the terrain, have long served as treetop detectors in remote sensing~\cite{popescu2002chm,jucker2017allometric}; we detect the peaks from the input scene, and associate the one whose allometric crown footprint covers the user's click as a ``free'' initial click.}
The main contributions of this work are as follows:
\begin{itemize}
    \item We propose \textbf{SelectAnyTree}, a promptable instance segmentation model for 3D forest LiDAR point clouds that enables users to select individual trees with a few clicks and interactively correct errors in dense canopies.
    
    \item The proposed SelectAnyTree operates with three lightweight stages, needing only 19.4 M parameters and $1.9$\,GB of GPU memory.
    Additionally, we exploit forest-aware information by using CHM treetops as free initial clicks.
    
    \item We establish a promptable benchmark for forest LiDAR under the SAM-style click-simulation protocol~\cite{kirillov2023segment,zhou2025pointsam,zhu2026partsam}, comparing promptable 3D baselines in the same setting across seven forest regions.
    SelectAnyTree reaches 79.9 Intersection-over-Union (IoU) from a single click, and every accuracy target with the fewest clicks.
\end{itemize}

%% file: sec/2_related-work.tex
\begin{figure*}[t]
\centering
\includegraphics[width=0.95\linewidth]{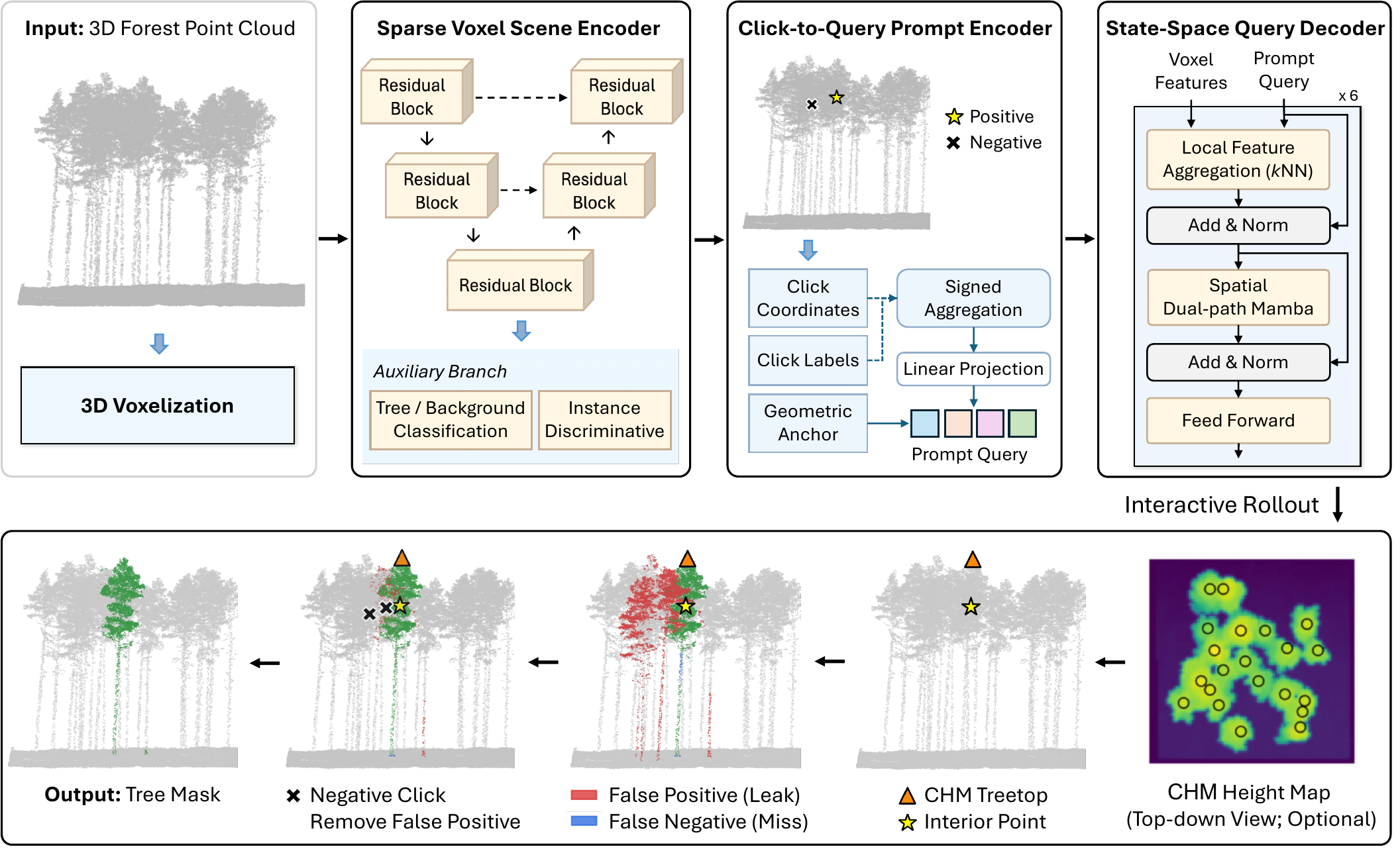}
\caption{\textbf{Overview of the proposed SelectAnyTree.} 
A forest point cloud is encoded once by the Sparse Voxel Scene Encoder into reusable voxel features.
Each click set is converted by the Click-to-Query Prompt Encoder into
a single content query that the State-Space Query Decoder turns into one tree mask.
Training and evaluation share the same Interactive Rollout.
The Canopy Height Model (CHM) treetop is a supporting, optional component that supplies a free, geometry-guided first prompt; the promptable core operates without it.}
\label{fig:architecture}
\vspace{-3mm}
\end{figure*}

\noindent \textbf{Forest point cloud instance segmentation.}
Delineating individual trees from LiDAR has a long history in remote sensing, where classical pipelines detect local maxima in a CHM and grow regions with watershed-style rules~\cite{naveed2019individual,popescu2002chm}.
These methods are effective for dominant, well-separated trees but degrade in dense, multi-layered canopies.
Deep learning has substantially improved robustness: ForAINet~\cite{xiang2024automated} couples a sparse convolutional U-Net~\cite{ronneberger2015u,graham2018spconv} with post-hoc clustering, and TreeLearn~\cite{henrich2024treelearn},   SegmentAnyTree~\cite{wielgosz2024segmentanytree,wielgosz2026segmentanytreev2} adapt modern point cloud backbones to class-agnostic tree segmentation, though many still rely on heuristic grouping.
More recent query-based models, including OneFormer3D~\cite{kolodiazhnyi2024oneformer3d}, ForestFormer3D~\cite{xiang2025forestformer3d}, and ForestMamba~\cite{nguyen_forestmamba}, formulate forest segmentation end-to-end and achieve good accuracy.
However, all these methods lack a way to select a specific tree or interactively correct an erroneous instance.

\vspace{5pt}
\noindent \textbf{3D point cloud segmentation backbones.}
Point-based networks such as PointNet~\cite{qi2017pointnet}, PointNet++~\cite{qi2017pointnetpp} and continuous-kernel convolutions like Kernel Point Convolution (KPConv)~\cite{thomas2019kpconv} pioneered direct learning on point sets, while sparse voxel convolutions~\cite{graham2018spconv,choy2019minkowski} enabled efficient large-scale scene processing, and test-time augmentation~\cite{vu2024testtime} further improves robustness.
Query-based set prediction, introduced by DEtection TRansformer (DETR)~\cite{carion2020detr} and extended to segmentation by Mask2Former~\cite{cheng2022masked} and OneFormer3D~\cite{kolodiazhnyi2024oneformer3d}, predicts instances as a set of learnable queries matched to ground truth.
The Point Transformer (PT) family~\cite{zhao2021point,wu2022ptv2} scales attention to large point clouds, and PTv3~\cite{wu2024ptv3} in particular serializes points along space-filling curves to trade precise locality for efficiency.
The quadratic cost of attention~\cite{vaswani2017attention} nonetheless scales poorly to large scenes, motivating a shift to linear-time state space models~\cite{gu2024mamba,liu2024vmamba}, which Local aggregation and State Space Models (LaSSM)~\cite{yao2026lassm} applies to 3D instance segmentation.
These backbones, however, are designed for automatic segmentation and do not support interactive, prompt-driven selection of individual instances.

\vspace{5pt}
\noindent \textbf{Promptable and interactive 3D segmentation.}
SAM~\cite{kirillov2023segment} established promptable segmentation in 2D, where clicks or boxes specify the target object.
In 3D, AGILE3D~\cite{yue2024agile3d} encodes user clicks as spatial-temporal queries and interactively segments multiple objects with a lightweight decoder.
{NPISeg3D~\cite{liu2025npising3d} further extends AGILE3D with a hierarchical neural process for uncertainty-aware mask refinement.}
Point-SAM~\cite{zhou2025pointsam} extends the SAM recipe to point clouds with a Transformer architecture and a 2D-to-3D label distillation engine.
PartSAM~\cite{zhu2026partsam} trains a promptable part segmentation model natively on large-scale 3D shapes.
These models target indoor objects, outdoor scenes, man-made shapes, or object parts, without explicitly accounting for the structural complexity of forest canopies. 
SelectAnyTree is therefore purpose-built for forest LiDAR scenes.

%% file: sec/3_proposed-method.tex
\noindent \textbf{Problem formulation.}
Given a 3D forest point cloud $\mathcal{P}=\{\mathbf{p}_i \in \mathbb{R}^{3}\}_{i=1}^{N}$ and a prompt consisting of a set of clicks $\mathcal{C}=\{(\mathbf{c}_j, \ell_j)\}_{j=1}^{P}$, where {$\mathbf{c}_j \in \mathcal{P}$ is a clicked point on the cloud} and $\ell_j \in \{0,1\}$ marks it as negative or positive, the goal of promptable tree segmentation is to predict a single binary mask $\hat{\mathbf{m}} \in \{0,1\}^{N}$ that segments the tree indicated by the prompt as:
\begin{equation}
g:\; (\mathcal{P}, \mathcal{C}) \;\mapsto\; \hat{\mathbf{m}}.
\end{equation}
This differs fundamentally from prior tree instance segmentation methods~\cite{xiang2025forestformer3d,nguyen_forestmamba}, which map $\mathcal{P}$ to a fixed set of all instance masks in one pass.
In the promptable setting, the user controls \emph{which} tree is segmented, and additional clicks iteratively refine the predicted mask.

\vspace{5pt}
\noindent \textbf{Method overview.}
The proposed SelectAnyTree is a promptable and interactive instance segmentation model that selects any individual tree from a 3D forest point cloud in a few consecutive clicks.
As shown in~\cref{fig:architecture}, it couples three lightweight stages, driven by an interactive rollout:
(1)~A forest point cloud is encoded \emph{once} into reusable voxel features by the Sparse Voxel Scene Encoder and shared with all prompts,
(2)~Each click set is turned into a single content query by the Click-to-Query Prompt Encoder,
(3)~The State-Space Query Decoder converts the prompt query into one voxel-resolution mask, mapped back to points,
(4)~An Interactive Rollout simulates a first prompt followed by correction clicks, and is used identically for training and evaluation.
Stages (1)--(3) form the promptable core that the model relies on in general use.
The CHM~\cite{popescu2002chm,jucker2017allometric} treetop of \cref{sec:chm_prompt} is an optional supporting component, supplying a first click for free where the canopy prior is reliable.

\subsection{Sparse Voxel Scene Encoder}
\label{sec:scene_encoding}
The point cloud is voxelized at resolution $v$ and processed by a sparse encoder~\cite{nguyen_forestmamba,gu2024mamba} to produce per-voxel features as:
\begin{equation}
\mathbf{F}=\{\mathbf{f}_n\}_{n=1}^{V}, \qquad \mathbf{f}_n \in \mathbb{R}^{C},
\end{equation}
where $V$ is the number of occupied voxels and $C$ is the backbone width.
We record the point-to-voxel assignment $\rho:\{1,\dots,N\}\!\to\!\{1,\dots,V\}$ and the voxel centroids $\{\mathbf{x}_n\}_{n=1}^{V}$, obtained by averaging the coordinates of the points falling in each voxel.
Encoding is performed once per scene and the features $\mathbf{F}$ are cached; segmenting many trees in the same scene then only repeats the lightweight prompt-encoding and decoding stages, an important property for interactive annotation.

An auxiliary branch attached to $\mathbf{F}$ predicts per-voxel tree/non-tree logits and instance-discriminative embeddings $\{\mathbf{e}_n\}$.
These are used to keep the backbone steady during promptable training (Sec.~\ref{sec:training}).

\subsection{Click-to-Query Prompt Encoder}
\label{sec:prompt_encoder}
The Prompt Encoder maps a click set $\mathcal{C}=\{(\mathbf{c}_j,\ell_j)\}_{j=1}^{P}$, with coordinates $\mathbf{c}_j\in\mathbb{R}^{3}$ and polarities $\ell_j\in\{0,1\}$, to a single content query $\mathbf{q}\in\mathbb{R}^{C}$ in the backbone-width space.
It fuses three ingredients: A geometric anchor $\bar{\mathbf{f}}$ sampled from the backbone, a positional code, and a label-aware aggregation of the clicks, combined as:
\begin{equation}
\label{eq:query_overview}
\begin{aligned}
\mathbf{q} &= \underbrace{\bar{\mathbf{f}}}_{\text{geometric anchor}}
\;+\; \mathbf{W}_o\,\underbrace{\frac{1}{P}\sum_{j=1}^{P}\operatorname{sgn}(\ell_j)\,\mathbf{t}_j}_{\text{label-aware aggregation}}, \\[5pt]
\mathbf{t}_j &= \mathrm{MLP}\big(
\underbrace{\gamma(\mathbf{c}_j)}_{\text{positional code}}
+ \underbrace{\mathbf{E}(\ell_j)}_{\text{polarity embedding}}\big),
\end{aligned}
\end{equation}
where $\mathbf{W}_o$ is a learned projection.

\vspace{4pt}
\noindent\textbf{Geometric anchor.}
Let {$\mathcal{C}^{+}=\{(\mathbf{c}_j,\ell_j)\in\mathcal{C}:\ell_j=1\}\subseteq\mathcal{C}$} be the set of positive clicks.
For each positive click we read the backbone feature of its \emph{nearest} voxel in the horizontal plane, and average these features over the positive clicks as:
\begin{equation}
\begin{aligned}
n_j^{\star} &= \operatorname*{arg\,min}_{n}\ \|\mathbf{x}_n^{xy} - \mathbf{c}_j^{xy}\|_2, \\
\bar{\mathbf{f}} &= \frac{1}{|\mathcal{C}^{+}|}\sum_{j:\,\ell_j=1}\mathbf{f}_{n_j^{\star}},
\end{aligned}
\end{equation}
where $\mathbf{x}_n^{xy}$ denotes the horizontal coordinate and $\mathbf{f}_n\in\mathbb{R}^{C}$ the backbone feature of voxel $n$.

\vspace{4pt}
\noindent\textbf{Positional and label encoding.}
Each click coordinate is normalized to $[-1,1]$ using the scene bounding box and embedded with a random-Fourier positional encoding~\cite{kirillov2023segment} as:
\begin{equation}
\gamma(\mathbf{c}) = \big[\sin(2\pi\, \mathbf{c}^{\!\top}\!\mathbf{B}),\;
\cos(2\pi\, \mathbf{c}^{\!\top}\!\mathbf{B})\big] \in \mathbb{R}^{2 d_{\mathrm{pe}}},
\end{equation}
where $\mathbf{B}\in\mathbb{R}^{3\times d_{\mathrm{pe}}}$ is a fixed Gaussian matrix.
A learned embedding $\mathbf{E}:\{0,1\}\to\mathbb{R}^{2 d_{\mathrm{pe}}}$ maps the click polarity $\ell$ (negative or positive) to $\mathbf{E}(\ell)$.
The two are summed and projected to backbone width by a lightweight Multi-Layer Perceptron (MLP), forming the per-click token $\mathbf{t}_j$ of Eq.~\eqref{eq:query_overview}.

\vspace{4pt}
\noindent\textbf{Signed aggregation.}
The signed weight in Eq.~\eqref{eq:query_overview} is:
\begin{equation}
\operatorname{sgn}(\ell)=
\begin{cases}
+1 & \ell=1 \ \text{(positive click)},\\
-1 & \ell=0 \ \text{(negative click)},
\end{cases}
\end{equation}
so positive clicks pull the query toward the target and negative clicks carve away neighbouring crowns, while averaging over the $P$ clicks keeps the scale stable as their number grows.
The projection $\mathbf{W}_o$, the MLP, and the polarity embedding $\mathbf{E}$ are learned jointly with the backbone and decoder under the promptable objective.
The query is further assigned a 3D anchor position equal to the mean of the positive click coordinates as:
\begin{equation}
\boldsymbol{\mu} = \frac{1}{|\mathcal{C}^{+}|}\sum_{j:\,\ell_j=1}\mathbf{c}_j,
\end{equation}
which the decoder uses for spatial aggregation.

\subsection{State-Space Query Decoder}
\label{sec:decoding}
The Query Decoder~\cite{nguyen_forestmamba,yao2026lassm} takes the query $\mathbf{q}$ and its anchor $\boldsymbol{\mu}$, together with the cached voxel features $\mathbf{F}$ and positions $\{\mathbf{x}_n\}$, and refines it through local $k$-nearest-neighbour aggregation and state-space modeling to predict voxel mask logits $\mathbf{M}$ and an objectness score $s$ as:
\begin{equation}
\mathbf{M} \in \mathbb{R}^{V}, \qquad s = \sigma(\hat{s}) \in [0,1].
\end{equation}
Across refinement rounds, a lightweight \emph{mask feedback} conditions each round on the previous mask, so that the round-$t$ voxel logit is given by:
\begin{equation}
\label{eq:mask_feedback}
M_v^{(t)} = \tilde{M}_v^{(t)} + \big(\mathbf{W}_h\mathbf{f}_v\big)^{\!\top}\psi\!\big(\sigma(M_v^{(t-1)})\big),
\end{equation}
with decoder logit $\tilde{M}_v^{(t)}$, low-rank projection $\mathbf{W}_h\in\mathbb{R}^{r\times C}$, and gate MLP $\psi$. The gate is zero-initialized, so it reduces to the feedback-free decoder at the start.
Many prompts are decoded together in one batched pass, and the voxel logits are lifted to points by the point-to-voxel assignment $\mathbf{M}^{\mathrm{pt}}_i=\mathbf{M}_{\rho(i)}$, giving $\hat{\mathbf{m}}=\mathbb{1}[\mathbf{M}^{\mathrm{pt}}>0]$.
The rollout runs at voxel resolution and only the final logits are lifted, so the click dynamics are identical at train and test.

\subsection{CHM-guided first prompt}
\label{sec:chm_prompt}
The user places a first positive click $\mathbf{c}_{\mathrm{in}}\in\mathbb{R}^{3}$ inside the target tree.
SelectAnyTree optionally adds a positive click at a treetop, obtained from the input point cloud alone: CHM~\cite{popescu2002chm,jucker2017allometric} and its maximum are computed \emph{before} any click and over the whole scene, so the target mask enters neither this detection nor the association below.
Rasterizing the points onto a horizontal grid and keeping the highest return per cell gives CHM, whose local maximum, taken at two resolutions and filtered by a minimum height $h_{\min}$ and a minimum peak separation $d_{\min}$, form a scene-level candidate set $\mathcal{T}=\{(\mathbf{c}^{(k)},h_k)\}_{k=1}^{K}$ of peaks and their heights, computed once and shared by all prompts.
Given $\mathbf{c}_{\mathrm{in}}$, we assign each peak an allometric crown radius $r_k{=}\alpha h_k^{\beta}$~\cite{jucker2017allometric} and prompt the nearest peak whose crown footprint covers the click as:
\begin{equation}
\label{eq:chm_assoc}
\mathbf{c}_{\mathrm{top}}=\mathbf{c}^{(k^{\star})},~~
k^{\star}=\operatorname*{arg\,min}_{k\,:\;\|\mathbf{c}^{(k),xy}-\mathbf{c}_{\mathrm{in}}^{xy}\|_2\,\le\, r_k}
\ \|\mathbf{c}^{(k),xy}-\mathbf{c}_{\mathrm{in}}^{xy}\|_2 .
\end{equation}
If no peak covers the click, as for a suppressed tree under a closed canopy, the prompt stays the interior click alone.
Since the rule reads only the click coordinate and the scene geometry, a wrong association is a failure mode the model must tolerate rather than an oracle it benefits from.

\subsection{Interactive rollout}
\label{sec:rollout}
The core of the proposed SelectAnyTree is a single click-simulation rollout used identically for training and evaluation, following the principle of SAM-style models~\cite{kirillov2023segment,zhou2025pointsam}.
Given the target ground-truth mask, the rollout proceeds for $T$ iterations as:
\begin{enumerate}
    \item \textbf{First prompt.} Initialize the click set per the {first-prompt} strategy of Sec.~\ref{sec:chm_prompt}.
    \item \textbf{Decode.} Build the query Eq.~\eqref{eq:query_overview} and decode the current mask $\hat{\mathbf{m}}^{(t)}$.
    \item \textbf{Correction.} Compare $\hat{\mathbf{m}}^{(t)}$ with the ground truth and sample one correction click from the larger error region: \emph{Positive} click in a missed (false-negative) region, or  \emph{Negative} click in a leaked (false-positive) region. Append it to the click set and repeat.
\end{enumerate}
Concretely, with $\mathbf{m}^{\star}$ the ground-truth mask, the false-negative and false-positive regions are $\mathcal{F}^{-}=\mathbf{m}^{\star}\wedge\neg\hat{\mathbf{m}}^{(t)}$ and $\mathcal{F}^{+}=\neg\mathbf{m}^{\star}\wedge\hat{\mathbf{m}}^{(t)}$, and the next click is placed at the most interior point of whichever region is larger, labelled positive for $\mathcal{F}^{-}$ and negative for $\mathcal{F}^{+}$, mirroring how a user corrects the model and how Point-SAM samples refinement prompts~\cite{zhou2025pointsam}.
This shared routine ensures the optimization objective matches the evaluation protocol exactly.

\subsection{Training objective}
\label{sec:training}
Promptable training runs the rollout of Sec.~\ref{sec:rollout} over the ground-truth trees in each scene and supervises every decoded mask.
The mask loss combines Binary Cross-Entropy (BCE) with a Dice term~\cite{milletari2016vnet} as:
\begin{equation}
\small
\mathcal{L}_{\mathrm{mask}}(\mathbf{M},\mathbf{m}^{\star}) =
\mathcal{L}_{\mathrm{BCE}}(\mathbf{M},\mathbf{m}^{\star}) +
\lambda_{\mathrm{Dice}}\,\mathcal{L}_{\mathrm{Dice}}(\sigma(\mathbf{M}),\mathbf{m}^{\star}),
\end{equation}
averaged over the simulated trees and over all rollout iterations, so every click count receives supervision.
Two auxiliary losses on the backbone are retained from the automatic model: Binary tree/non-tree cross-entropy on the voxel semantics, and Discriminative embedding loss~\cite{de2017semantic} that pulls together voxels of the same tree and pushes apart different trees.
The total objective is defined as:
\begin{equation}
\mathcal{L} = \lambda_{\mathrm{p}}\,\mathcal{L}_{\mathrm{mask}}
+ \lambda_{\mathrm{bin}}\,\mathcal{L}_{\mathrm{bin}}
+ \lambda_{\mathrm{dis}}\,\mathcal{L}_{\mathrm{dis}},
\end{equation}
with scalar weights $\lambda_{*}$.
The full specification of each loss term is provided in the Supplementary Material.

%% file: sec/4_evaluation.tex
\begin{table*}[t]
\centering
\caption{Interactive tree instance segmentation.
For the in-distribution evaluation, we use the FOR-instanceV2~\cite{xiang2025forestformer3d} test set, whose trees are unseen but come from the regions used for training.
For cross-dataset generalization, we evaluate on the LAUTx dataset~\cite{tockner2022lautx}, a separate dataset excluded from training entirely.
The last column summarizes whether the CHM prompt improves IoU: \cmark{} the CHM prompt improves IoU, \xmark{} it never improves IoU.
Best and second-best results per column are in \textbf{bold} and \underline{underlined}, respectively.}
{%
\begin{tabular}{l|ccc|ccc|ccc|ccc|c}
\toprule
\multirowcell{3}[-4pt][l]{{Method}}
& \multicolumn{6}{c|}{In-distribution}
& \multicolumn{6}{c|}{Cross-dataset Generalization}
& \multirowcell{3}[-4pt]{CHM\\helps?} \\
\cmidrule(lr){2-7} \cmidrule(lr){8-13}
& \multicolumn{3}{c|}{IoU @ Clicks $\uparrow$} & \multicolumn{3}{c|}{NoC @ IoU $\downarrow$}
& \multicolumn{3}{c|}{IoU @ Clicks $\uparrow$} & \multicolumn{3}{c|}{NoC @ IoU $\downarrow$} & \\
\cmidrule(lr){2-4} \cmidrule(lr){5-7} \cmidrule(lr){8-10} \cmidrule(lr){11-13}
& 1 & 3 & 5 & 0.7 & 0.8 & 0.9
& 1 & 3 & 5 & 0.7 & 0.8 & 0.9 & \\
\midrule
AGILE3D~\cite{yue2024agile3d}
& 43.8 & 64.6 & 71.9 & 4.6 & 5.9 & 7.8
& 40.8 & 62.0 & 68.7 & 5.2 & 6.4 & 8.3 & \multirow{2}{*}{\xmark} \\
\hspace{5pt}+ w/ CHM
& 43.7 & 59.3 & 66.7 & 5.3 & 6.5 & 8.3
& 40.6 & 56.0 & 62.5 & 5.7 & 7.0 & 8.8 & \\

\cmidrule(lr){1-14}
NPISeg3D~\cite{liu2025npising3d}
& {36.9} & {61.0} & {68.5} & {5.0} & {6.3} & {8.1}
& {38.3} & {59.3} & {65.1} & {5.3} & {6.9} & {8.8} & \multirow{2}{*}{\cmark} \\
\hspace{5pt}+ w/ CHM
& 46.4 & 63.9 & 68.3 & 4.8 & 6.2 & 8.2
& 42.6 & 61.7 & 69.3 & 5.1 & 6.5 & 8.5 & \\ 

\cmidrule(lr){1-14}
Point-SAM~\cite{zhou2025pointsam}
& 53.4 & 74.8 & 80.6 & {3.4} & {4.4} & 6.1
& 54.5 & 75.0 & 81.2 & 3.3 & {4.5} & 6.5 & \multirow{2}{*}{\cmark} \\
\hspace{5pt}+ w/ CHM
& 55.2 & 75.1 & \underline{81.6} & {3.4} & {4.4} & {6.0}
& 55.7 & 76.2 & 81.5 & 3.2 & {4.5} & 6.5 & \\ 

\cmidrule(lr){1-14}
PartSAM~\cite{zhu2026partsam}
& 48.8 & 71.2 & 78.3 & 3.7 & 4.8 & 6.6
& 31.4 & 53.9 & 63.9 & 5.9 & 7.1 & 8.5 & \multirow{2}{*}{\cmark} \\
\hspace{5pt}+ w/ CHM
& 53.1 & 74.0 & 80.8 & 3.6 & 4.7 & 6.8
& 44.8 & 66.1 & 75.5 & 4.3 & 5.8 & 7.8 & \\



\cmidrule(lr){1-14}
{\textbf{SelectAnyTree (Prop.)}}
& \underline{77.9} & \underline{81.3} & 81.1 & \underline{3.0} & \underline{3.9} & \underline{5.5}
& \underline{76.5} & \underline{80.3} & \underline{81.6} & \underline{2.7} & \underline{3.7} & \underline{6.0} & \multirow{2}{*}{\cmark} \\
\hspace{5pt}+ w/ CHM
& \textbf{79.9} & \textbf{81.9} & \textbf{82.4} & \textbf{2.9} & \textbf{3.8} & \textbf{5.4}
& \textbf{77.7} & \textbf{81.5} & \textbf{82.2} & \textbf{2.6} & \textbf{3.4} & \textbf{5.7} \\

\bottomrule
\end{tabular}%
}
\label{table:tree_instance_generalization}
\end{table*}


\subsection{Experimental conditions}
\noindent\textbf{Dataset.}
Following ForestFormer3D~\cite{xiang2025forestformer3d} and ForestMamba~\cite{nguyen_forestmamba}, we evaluate on the FOR-instanceV2 benchmark~\cite{xiang2025forestformer3d}, which spans seven geographically diverse forest regions across four continents and covers boreal, temperate, and tropical forest types: CULS (Czech Republic), NIBIO (Norway), RMIT (Australia), SCION (New Zealand), TUWIEN (Austria), BlueCat (Czech Republic), and YuChen (French Guiana).
The dataset provides point-level instance annotations for $11{,}134$ tree instances, split into $8{,}016$ for training, $1{,}374$ for validation, and $1{,}744$ for testing.
To assess cross-dataset generalization, we additionally evaluate on the LAUTx dataset~\cite{tockner2022lautx}, a set of $516$ individual-tree point clouds from Austrian forest inventory plots that is held out entirely from training.
Only the raw $(x,y,z)$ coordinates are used as input, without color or intensity.
Each plot is cropped into cylindrical patches of radius $16$\,m, voxelized at $0.2$\,m, and subsampled to at most $640$k points per patch.

\vspace{5pt}
\noindent\textbf{Evaluation protocol.}
To benchmark the interactive setting without a human in the loop, we simulate the ``user'' from the ground-truth mask using the same rollout employed during training~\cite{zhu2026partsam,zhou2025pointsam}.
The first click is placed deterministically at the in-mask point closest to the mask centroid, and each later click at the most interior point of the larger error region.
The full specification is given in the Supplementary Material.
We report two complementary families of model performance metrics: 
First, for the \emph{interactive} setting, we report the IoU reached after $1$, $3$, and $5$ user clicks per instance, and the Number of Clicks (NoC) required to reach IoU targets of $0.70$, $0.80$, and $0.90$ (lower is better), using an evaluation horizon of up to $10$ clicks.
Second, for comparison with task-specific methods, we report \emph{whole-scene} F$_1$ at an IoU threshold of $0.50$ as a function of the click budget, where $k\!=\!0$ prompts are sampled automatically on a regular grid over the scene and $k\!\ge\!1$ gives each tree $k$ simulated clicks.


\vspace{5pt}
\noindent\textbf{Compared methods.}
We compare SelectAnyTree against two groups of methods.
\emph{Promptable} segmentation methods, namely AGILE3D~\cite{yue2024agile3d}, NPISeg3D~\cite{liu2025npising3d}, Point-SAM~\cite{zhou2025pointsam}, and PartSAM~\cite{zhu2026partsam}, are re-trained on the same training data as SelectAnyTree.
\emph{Task-specific} segmentation methods, namely TreeLearn~\cite{henrich2024treelearn}, ForAINet~\cite{xiang2024automated}, OneFormer3D~\cite{kolodiazhnyi2024oneformer3d}, ForestFormer3D~\cite{xiang2025forestformer3d}, and ForestMamba~\cite{nguyen_forestmamba}, predict the entire scene without user input.

\begin{table*}[t]
\centering
\caption{Model size and inference efficiency on the {FOR-instanceV2}~\cite{xiang2025forestformer3d} test set.
Time [ms] is mean$\,\pm\,$std GPU wall-clock time per scene for a $k$-click interactive session.
All methods are timed on the same input scenes and the same trees.
Best and second-best results per column are in \textbf{bold} and
\underline{underlined}, respectively.}
{
\begin{tabular}{l|c|ccc|c}
\toprule
\multirowcell{2}[-2pt][l]{{Method}} &
\multirowcell{2}[-2pt][c]{Params\\{[M]} $\downarrow$} &
\multicolumn{3}{c|}{Time [ms] @ Clicks $\downarrow$} &
\multirowcell{2}[-2pt][c]{Peak GPU\\{[GB] $\downarrow$}} \\
\cmidrule(lr){3-5}
& & 1 & 3 & 5 & \\
\midrule
AGILE3D~\cite{yue2024agile3d}     & \hspace{2pt} \underline{39.3} & \hspace{5pt} 412\,$\pm$\,241  & \hspace{-10pt} ~~~1,828\,$\pm$\,~~~653  & 3,286\,$\pm$\,1,043  & \hspace{2pt} \textbf{1.0} \\
NPISeg3D~\cite{liu2025npising3d}     & \hspace{2pt} {40.0} & \hspace{5pt} {249\,$\pm$\,131}  & \hspace{-10pt} {~~~1,307\,$\pm$\,~~~386}  & \hspace{-10pt} {~~~2,353\,$\pm$\,~~~664}  & \hspace{2pt} {3.1} \\
Point-SAM~\cite{zhou2025pointsam} &              311.0            & 1,612\,$\pm$\,602 & 5,062\,$\pm$\,2,035 & 8,587\,$\pm$\,3,472 & 22.4 \\
PartSAM~\cite{zhu2026partsam}     &              225.0            & 1,549\,$\pm$\,581 & \hspace{-10pt} ~~~2,558\,$\pm$\,~~~986  & 3,562\,$\pm$\,1,423  & \hspace{2pt} 2.4 \\
\cmidrule(lr){1-6}
\textbf{SelectAnyTree (Prop.)}              & \hspace{2pt} \textbf{19.4}   &  \hspace{4pt} \textbf{122\,$\pm$\,\hspace{4pt}98}      & ~~~\underline{298\,$\pm$\,~~~196} & ~~~\underline{571\,$\pm$\,~~~184}  & \hspace{2pt} \underline{1.9} \\
\hspace{5pt}+ w/ CHM            & \hspace{2pt} \textbf{19.4}   & \hspace{5pt} \underline{164\,$\pm$\,159}  & ~~~\textbf{280\,$\pm$\,~~~156}    & ~~~\textbf{549\,$\pm$\,~~~126}    & \hspace{2pt} \underline{1.9} \\
\bottomrule
\end{tabular}
}
\label{table:model_efficiency}
\vspace{-2mm}
\end{table*}

\vspace{5pt}
\noindent\textbf{Implementation details.}
SelectAnyTree pairs a five-level Sparse Voxel Scene Encoder (width $32$, $d_{\mathrm{state}}{=}16$, slab thickness $5$, bidirectional scanning) with a State-Space Query Decoder ($L{=}6$, model dimension $256$, state dimension $64$).
The Click-to-Query Prompt Encoder uses a random-Fourier positional encoding ($d_{\mathrm{pe}}{=}64$) and reads the geometric anchor from the nearest voxel, trained jointly with the backbone.
CHM treetop detection uses grid resolutions $0.3$/$0.7$\,m, minimum tree height $h_{\min}{=}1.5$\,m, and minimum peak separation $d_{\min}{=}1.0$\,m; the click-to-treetop association of Eq.~\eqref{eq:chm_assoc} uses the allometric crown radius $r_k{=}\alpha h_k^{\beta}$ with $(\alpha,\beta){=}(0.25,0.5)$. 
We train the click-simulation rollout with five prompt iterations per tree, a Dice weight of $2.0$, and up to $128$ trees per scene, using AdamW~\cite{loshchilov2019decoupled} ($\mathrm{lr}{=}10^{-4}$, weight decay $0.05$) with a polynomial schedule and gradient clipping, for $1{,}000$ epochs on a single NVIDIA A6000 GPU.
The \mbox{+ w/ CHM} setting adds the automatic treetop as a free positive click.

\subsection{Quantitative results}
\noindent\textbf{Interactive tree instance segmentation.}
\Cref{table:tree_instance_generalization} reports IoU at increasing click budgets and NoC at increasing IoU targets, in-distribution on FOR-instanceV2~\cite{xiang2025forestformer3d} and cross-dataset on the unseen LAUTx dataset~\cite{tockner2022lautx}.
The gap was clearest at the first prompt, where a single click brought SelectAnyTree to $79.9$ IoU in-distribution, $24.7$ points above the strongest baseline, Point-SAM~\cite{zhou2025pointsam}.
This was carried by the promptable core rather than the canopy prior: without CHM it already reached $77.9$, and the treetop added $2.0$ points.
Consequently, SelectAnyTree hit every IoU target with the fewest clicks, while Point-SAM and PartSAM~\cite{zhu2026partsam} needed about five clicks to match its single-click IoU.
Additionally, we run a bootstrap analysis for SelectAnyTree that confirmed these gains were statistically stable, with standard deviations below $0.65$ IoU and $0.13$ NoC.

On the unseen LAUTx dataset, SelectAnyTree retained $77.7$ IoU from a single click, $22.0$ points above the strongest baseline, again reaching every IoU target with the fewest clicks.
The CHM prompt, a consistent and cost-free geometric cue, also helped under this domain shift.
The lead narrowed as clicks accumulated, with Point-SAM within $0.7$ IoU at five clicks, showing that SelectAnyTree's advantage was concentrated in the low-click regime.

\begin{table}[t]
\centering
\caption{
Ablation of the query prompt and mask feedback on the FOR-instanceV2~\cite{xiang2025forestformer3d}.
All variants use the + w/ CHM setting.
The query fuses a geometric anchor (Feat.), a positional code of the click coordinate (Pos.), and a polarity embedding (Pol.).
}
\setlength{\tabcolsep}{3.5pt}
\resizebox{1.0\linewidth}{!}
{
\begin{tabular}{lccc|ccc}
\toprule
\multirowcell{2}[-2pt][l]{Variant}
& \multicolumn{3}{c|}{Query terms}
& \multicolumn{3}{c}{IoU @ Clicks $\uparrow$} \\
\cmidrule(lr){2-4} \cmidrule(lr){5-7}
& Feat. & Pos. & Pol. & 1 & 3 & 5 \\
\midrule
\textbf{SelectAnyTree (Prop.)}   & \cmark & \cmark & \cmark & {79.9} & {81.9} & {82.4} \\
\midrule
\rowcolor[HTML]{E5E5E5}
\multicolumn{7}{l}{\textit{Query composition}} \\ \midrule
Geometric anchor only              & \cmark & \xmark & \xmark & 79.2 & 80.1 & 80.0 \\
Click token only                 & \xmark & \cmark & \cmark & 77.8 & 80.9 & 81.0 \\
\midrule
\rowcolor[HTML]{E5E5E5}
\multicolumn{7}{l}{\textit{Local feature pooling (nearest voxel $\rightarrow$ cylinder)}} \\ \midrule
Cylinder, $r_p{=}0.5$\,m         & \multicolumn{3}{c|}{--} & 78.9 & 81.2 & 81.6 \\
Cylinder, $r_p{=}1.5$\,m          & \multicolumn{3}{c|}{--} & 79.6 & 81.9 & 82.1 \\
\midrule
\rowcolor[HTML]{E5E5E5}
\multicolumn{7}{l}{\textit{Mask feedback}} \\ \midrule
w/o mask feedback                & \multicolumn{3}{c|}{--} & 78.2 & 80.6 & 81.2 \\
\bottomrule
\end{tabular}
}
\label{table:ablation_prompt_encoder}
\vspace{-3mm}
\end{table}

\vspace{5pt}
\noindent\textbf{Efficiency.}
\Cref{table:model_efficiency} reports model size, inference time, and peak GPU memory on the {FOR-instanceV2} test set.
The proposed SelectAnyTree was the most compact model at {19.4\,M} parameters, up to {16$\times$} fewer than baselines, yet achieved the fastest inference at every click budget.
At 5 clicks, it completed an interactive session in 571\,ms, over {4$\times$} faster than NPISeg3D~\cite{liu2025npising3d}, the fastest promptable baseline, and {15$\times$} faster than Point-SAM.
This efficiency stems from encoding the scene once and decoding all prompts in one batched pass, so the cost of selecting an additional tree is a prompt encoding and a decode rather than a re-encoding of the scene.
SelectAnyTree also used only {1.9\,GB} of GPU memory, {12$\times$} less than Point-SAM.
Additionally, the \mbox{+ w/ CHM} setting had a comparable inference cost, since treetops are detected only once per scene.

\vspace{5pt}
\noindent\textbf{Ablation study.}
\Cref{table:ablation_prompt_encoder} isolates the three parts of the promptable design on the FOR-instanceV2~\cite{xiang2025forestformer3d} test set.
\begin{itemize}[leftmargin=*,itemsep=1pt,topsep=2pt]
\item \emph{Query composition.} The anchor and click token were complementary: the anchor alone gave an accurate first click but plateaued without cues to steer corrections, while the click token alone started lower yet refined better; combining them performed best.
\item \emph{Local feature pooling.} The nearest-voxel anchor was best or tied at every click budget; a larger cylinder radius recovered part of the gap, but averaging blurred the anchor and leaked across adjacent crowns.
\item \emph{Mask feedback.} Conditioning each round on the previous mask added $1.2$--$1.7$ IoU, improving iterative refinement.
\end{itemize}
In short, the anchor supplied content, the click token steering, and the mask feedback state across rounds.

\begin{table}[t]
\centering
\caption{Whole-scene tree instance segmentation on the
FOR-instanceV2~\cite{xiang2025forestformer3d} test set, reported at $k$ clicks per tree.
Task-specific methods predict every instance without user input and therefore occupy only the $k\!=\!0$ column.
For promptable methods, $k\!=\!0$ uses grid-sampled prompts and $k\!\ge\!1$ gives each tree $k$ simulated user clicks.
F$_1$ is computed at IoU $=0.5$.
Best and second-best results per group are in \textbf{bold} and
\underline{underlined}, respectively.}
\resizebox{0.95\linewidth}{!}{
\begin{tabular}{l|cccc}
\toprule
\multirowcell{2}[-2pt][l]{{Method}}
    & \multicolumn{4}{c}{F$_1$ @ $k$ clicks $\uparrow$} \\
\cmidrule(lr){2-5}
    & 0 & 1 & 3 & 5 \\
\midrule
\rowcolor[HTML]{E5E5E5}
\multicolumn{5}{l}{\textit{Task-specific}} \\ \midrule
OneFormer3D~\cite{kolodiazhnyi2024oneformer3d}  & 74.0 & -- & -- & -- \\
TreeLearn~\cite{henrich2024treelearn}           & 50.6 & -- & -- & -- \\
ForAINet~\cite{xiang2024automated}              & 72.8 & -- & -- & -- \\
ForestFormer3D~\cite{xiang2025forestformer3d}   & \underline{82.3} & -- & -- & -- \\
ForestMamba~\cite{nguyen_forestmamba}           & \textbf{83.4} & -- & -- & -- \\
\midrule
\rowcolor[HTML]{E5E5E5}
\multicolumn{5}{l}{\textit{Promptable}} \\ \midrule
AGILE3D~\cite{yue2024agile3d}     & 14.2 & 45.4 & 76.1 & 83.3 \\
NPISeg3D~\cite{liu2025npising3d}  & 15.7 & 40.2 & 74.7 & 81.9 \\
Point-SAM~\cite{zhou2025pointsam} & 35.9 & 49.4 & 81.0 & 88.5 \\
PartSAM~\cite{zhu2026partsam}     & 35.8 & 52.3 & 85.0 & 88.0 \\
\cmidrule(lr){1-5}
\textbf{SelectAnyTree (Prop.)}           & \underline{38.7} & \underline{86.1} & \underline{89.0} & \underline{89.1} \\
\hspace{5pt}+ w/ CHM              & \textbf{41.7} & \textbf{89.0} & \textbf{90.2} & \textbf{90.6} \\
\bottomrule
\end{tabular}
}
\label{table:tree_instance}
\vspace{-3mm}
\end{table}

\vspace{5pt}
\noindent\textbf{Whole-scene tree instance segmentation.}
\Cref{table:tree_instance} compares promptable methods against task-specific automatic models as the click budget grows.
At $k\!=\!0$, with grid-sampled rather than user-placed prompts, every promptable method trailed the automatic models: SelectAnyTree led its group at $41.7$ F$_1$, over five points above Point-SAM and PartSAM but far below ForestMamba~\cite{nguyen_forestmamba} at $83.4$. 
Here, the CHM prompt mattered most, adding $3.0$ F$_1$ by seeding one clean query per treetop.
A single simulated click per tree was sufficient to overturn this: SelectAnyTree reached $89.0$ F$_1$, exceeding the strongest task-specific model by $5.6$ points, while PartSAM required three clicks, Point-SAM five, and AGILE3D and NPISeg3D remained just short of ForestMamba even at five.
The families also differ in what they leave an annotator: A task-specific model returns a fixed mask set, where a wrong instance can be discarded but not repaired, whereas a promptable model spends clicks exactly where the prediction fails.

\begin{figure*}[t]
\centering
\begin{subfigure}{0.95\linewidth}
\centering
\includegraphics[width=\textwidth]{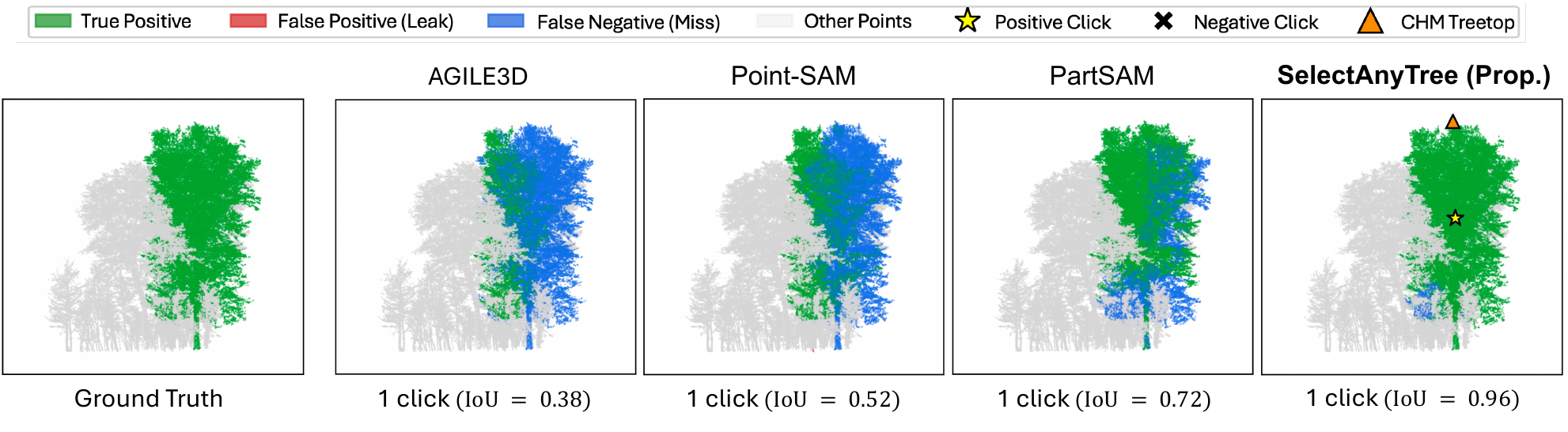}
\subcaption{BlueCat (Czech Republic)~\cite{xiang2025forestformer3d}: A dense temperate forest with strongly overlapping crowns.}
\label{fig:qualitative_bluecat}
\end{subfigure}
\begin{subfigure}{0.95\linewidth}
\centering
\includegraphics[width=\textwidth]{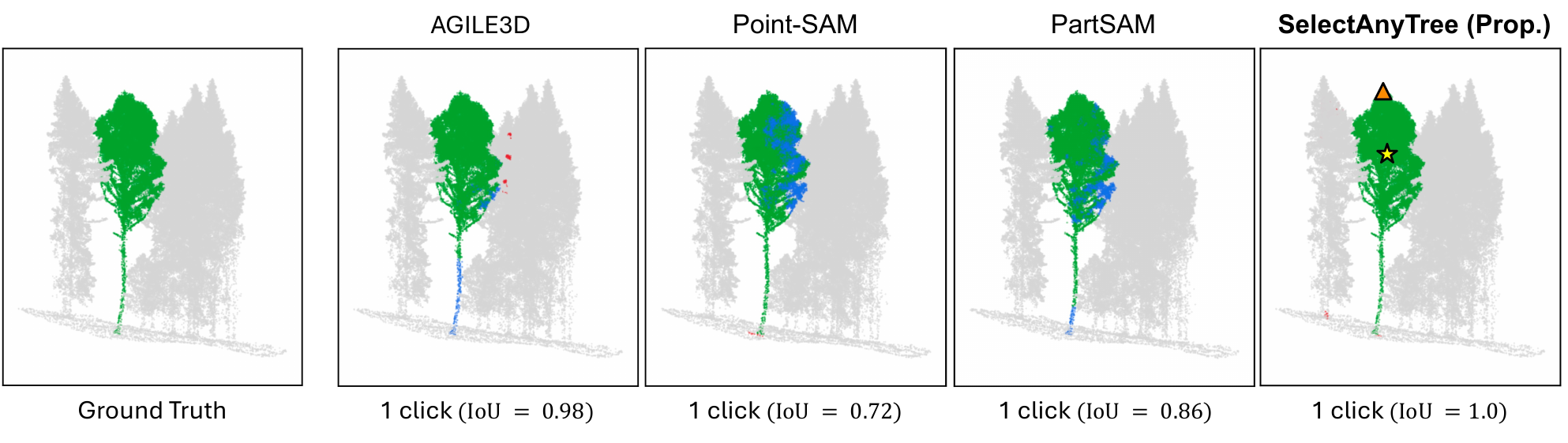}
\subcaption{NIBIO (Norway)~\cite{xiang2025forestformer3d}: A boreal conifer plot.}
\label{fig:qualitative_nibio}
\end{subfigure}
\caption{Single-click tree segmentation.
Comparison of the three strongest promptable baselines:
AGILE3D~\cite{yue2024agile3d}, Point-SAM~\cite{zhou2025pointsam}, and
PartSAM~\cite{zhu2026partsam}, against the proposed SelectAnyTree.}
\label{fig:qualitative}
\vspace{-3mm}
\end{figure*}

\subsection{Qualitative results}
\Cref{fig:qualitative} presents a single-click comparison on two contrasting regions, the BlueCat (Czech Republic) and NIBIO (Norway) test scenes. 
In the densely entangled BlueCat canopy, the promptable baselines either leaked into adjacent crowns (red) or left large portions of the target unsegmented (blue), yielding only $0.38$--$0.72$ IoU from a single click.
In contrast, the proposed SelectAnyTree delineated the target tree almost completely; the CHM treetop (orange triangle) anchored the canopy apex and the interior click grounded the trunk, two complementary cues that resolve the ambiguity a lone click leaves in overlapping foliage.
On the NIBIO boreal conifer scenes, the methods showed closer results. 
Yet, only the proposed SelectAnyTree recovered the tree exactly from one prompt.
Additional qualitative results, including top--down and side views, are provided in the Supplementary Material.

%% file: sec/5_conclusion.tex
We proposed \textbf{SelectAnyTree}, a promptable instance segmentation model that brings the click-driven ``segment anything'' paradigm to 3D forest LiDAR point clouds.
Unlike prior tree instance segmentation methods that segment all trees in a scene automatically in a single pass, it lets a user select and refine any individual tree in a few clicks.
It couples three lightweight stages: (1) Sparse Voxel Scene Encoder that embeds the point cloud \emph{once} into features reused by every prompt, (2) Click-to-Query Prompt Encoder that turns each click into a single content query, and (3) State-Space Query Decoder that converts it into one tree mask in linear time, with a mask feedback that conditions each refinement round on the previous mask.
All three stages are trained jointly under a click-simulation rollout used identically in training and evaluation, specializing the model for forest structure.
SelectAnyTree required just 19.4 M parameters, the fewest among all compared promptable models, and used only 1.9 GB of GPU memory, enabling the fastest inference and a real-time user experience with up to 164 ms per click.
Additionally, we exploit forest-aware information by using CHM treetops as a free initial click, which helps where the canopy prior is reliable and can be disabled where it is not.
Experimentally, SelectAnyTree reached 79.9 IoU from a single click, 24.7 points above the strongest promptable baseline, and achieved the lowest number of clicks to every IoU target, while remaining the most compact and efficient model across both in-distribution and cross-dataset evaluations.

\vspace{5pt}
\noindent\textbf{Limitations.}
All clicks are simulated from ground-truth masks, as in prior promptable segmentation models~\cite{yue2024agile3d,zhou2025pointsam}.
This ensures reproducible comparisons but idealizes human interaction, since real annotators click imprecisely and without access to the ground-truth error map.
The first click is also favourable by construction, so the reported single-click accuracies represent a best-case setting.

\vspace{5pt}
\noindent\textbf{Future work.}
We plan to evaluate SelectAnyTree with human annotators, measuring click behaviour, per-click accuracy, and annotation time.
We also aim to explore richer prompts, large-scale promptable pre-training across diverse forests, and integration into annotation tools for downstream tasks, such as \url{https://3dtrees.earth/}.

%% file: sec/6_appendix.tex
\section{Promptable Training Loss}
\label{sec:appendix_loss}

SelectAnyTree is trained end-to-end under the promptable objective as:
\begin{equation}
\mathcal{L} =
\lambda_{\mathrm{p}}\,\mathcal{L}_{\mathrm{mask}}
+ \lambda_{\mathrm{bin}}\,\mathcal{L}_{\mathrm{bin}}
+ \lambda_{\mathrm{dis}}\,\mathcal{L}_{\mathrm{dis}},
\end{equation}
with $\lambda_{\mathrm{p}}{=}\lambda_{\mathrm{bin}}{=}\lambda_{\mathrm{dis}}{=}1.0$.
$\mathcal{L}_{\mathrm{mask}}$ is the promptable mask loss applied to every decoded prompt, while $\mathcal{L}_{\mathrm{bin}}$ and $\mathcal{L}_{\mathrm{dis}}$ are auxiliary losses that keep the shared backbone discriminative.

\vspace{4pt}
\noindent\textbf{Promptable mask loss $\mathcal{L}_{\mathrm{mask}}$.}
For a prompted tree with voxel mask logits $\mathbf{M}\in\mathbb{R}^{V}$ and binary ground-truth mask $\mathbf{m}^{\star}\in\{0,1\}^{V}$, the loss combines binary cross-entropy with a Dice term~\cite{milletari2016vnet} as follows:
\begin{equation}
\mathcal{L}_{\mathrm{mask}} = \mathcal{L}_{\mathrm{BCE}}(\mathbf{M},\mathbf{m}^{\star})
+ \lambda_{\mathrm{Dice}}\,\mathcal{L}_{\mathrm{Dice}}(\sigma(\mathbf{M}),\mathbf{m}^{\star}),
\end{equation}
with $\lambda_{\mathrm{dice}}{=}2.0$ and $\sigma(\cdot)$ the sigmoid. Writing $p_n{=}\sigma(\mathbf{M}_n)$,
\begin{align}
\mathcal{L}_{\mathrm{BCE}}
&= -\frac{1}{V}\sum_{n=1}^{V}\!\Big[m^{\star}_n\log p_n + (1{-}m^{\star}_n)\log(1{-}p_n)\Big],\\
\mathcal{L}_{\mathrm{Dice}}
&= 1 - \frac{2\sum_{n} p_n\,m^{\star}_n + \epsilon}{\sum_{n} p_n + \sum_{n} m^{\star}_n + \epsilon},
\end{align}
with stabilizer $\epsilon{=}10^{-3}$.
The Dice term is scale-invariant to the number of foreground voxels, compensating for the large background imbalance of a single tree within a forest patch.
Crucially, $\mathcal{L}_{\mathrm{mask}}$ is evaluated at \emph{every} iteration of the interactive rollout (Sec.~\ref{sec:appendix_clicksim}) and averaged over the prompted trees and the rollout steps, so the model is supervised at all click counts it is scored with.

\vspace{4pt}
\noindent\textbf{Auxiliary tree/non-tree loss $\mathcal{L}_{\mathrm{bin}}$.}
An auxiliary head predicts a per-voxel tree/non-tree log-probability. With binary target $t_n\in\{0,1\}$ ($t_n{=}1$ if voxel $n$ belongs to any tree), $\mathcal{L}_{\mathrm{bin}}$ is the cross-entropy over the two classes. This keeps the backbone's foreground prediction reliable, which the CHM treetop prompt depends on.

\vspace{4pt}
\noindent\textbf{Auxiliary discriminative loss $\mathcal{L}_{\mathrm{dis}}$.}
Following~\cite{de2017semantic}, a discriminative embedding loss is applied to the tree-voxel embeddings $\{\mathbf{e}_n\}$ to pull voxels of the same tree toward their instance centroid and push different instances apart, with intra-/inter-instance margins $(\delta_v,\delta_d){=}(0.5,1.5)$. This shapes an instance-aware feature space that benefits both the automatic seeding path and the promptable queries.

\begin{table*}[t]
\centering
\caption{Summary of the FOR-instanceV2 dataset~\cite{xiang2025forestformer3d}.
\#Train, \#Val, and \#Test denote the number of annotated tree instances per split.
Sensors are Unmanned (ULS), Terrestrial (TLS), and Mobile (MLS) Laser Scanning.}
\label{tab:appendix_dataset}
{%
\begin{tabular}{llllrrr}
\toprule
Region & Country & Forest type & Sensor & \#Train & \#Val & \#Test \\
\midrule
CULS    & Czech Republic    & Temperate coniferous & ULS       & 6    & 21  & 20   \\
NIBIO   & Norway        & Boreal coniferous    & ULS / MLS & 2,401 & 572 & 1,021 \\
RMIT    & Australia     & Dry eucalypt         & ULS       & 159  & 0   & 64   \\
SCION   & New Zealand   & Temperate coniferous & ULS       & 69   & 23  & 43   \\
TUWIEN  & Austria       & Temperate deciduous  & ULS       & 115  & 0   & 35   \\
BlueCat & Czech Republic    & Temperate mixed      & TLS       & 5,032 & 735 & 537  \\
YuChen  & French Guiana & Tropical             & ULS       & 234  & 23  & 24   \\
\midrule
Total   &               &                      &           & 8,016 & 1,374 & 1,744 \\
\bottomrule
\end{tabular}%
}
\end{table*}

\begin{algorithm}[t]
\caption{Interactive click-simulation rollout (shared by training and evaluation).}
\label{alg:clicksim}
\begin{algorithmic}[1]
\Require Scene features; target GT mask $\mathbf{m}^{\star}$; iterations $T$; CHM mode.
\Ensure Per-iteration predicted masks $\{\hat{\mathbf{m}}^{(t)}\}_{t=1}^{T}$.
\State $\mathcal{C} \gets \{(\mathbf{c}_{\mathrm{in}}, +1)\}$ \Comment{interior click (nearest to centroid)}
\If{CHM mode}
  \State $\mathcal{C} \gets \mathcal{C} \cup \{(\mathbf{c}_{\mathrm{top}}, +1)\}$ \Comment{nearest CHM peak whose crown footprint covers $\mathbf{c}_{\mathrm{in}}$}
\EndIf
\For{$t = 1$ to $T$}
  \State $\mathbf{q} \gets \textsc{PromptEncoder}(\mathcal{C})$ \Comment{Sec.~3.2}
  \State $\hat{\mathbf{m}}^{(t)} \gets \textsc{Decode}(\mathbf{q})$ \Comment{Sec.~3.3}
  \If{$t < T$}
    \State $\mathcal{F}^{-} \gets \mathbf{m}^{\star}\wedge\neg\hat{\mathbf{m}}^{(t)}$;\quad
           $\mathcal{F}^{+} \gets \neg\mathbf{m}^{\star}\wedge\hat{\mathbf{m}}^{(t)}$
    \If{$|\mathcal{F}^{-}| \ge |\mathcal{F}^{+}|$ \textbf{and} $|\mathcal{F}^{-}| > 0$}
      \State $\mathbf{c} \gets$ most interior point of $\mathcal{F}^{-}$;\quad
             $\mathcal{C} \gets \mathcal{C} \cup \{(\mathbf{c}, +1)\}$
    \ElsIf{$|\mathcal{F}^{+}| > 0$}
      \State $\mathbf{c} \gets$ most interior point of $\mathcal{F}^{+}$;\quad
             $\mathcal{C} \gets \mathcal{C} \cup \{(\mathbf{c}, 0)\}$
    \EndIf
  \EndIf
\EndFor
\State \Return $\{\hat{\mathbf{m}}^{(t)}\}_{t=1}^{T}$
\end{algorithmic}
\end{algorithm}

\section{Click-simulation Protocol}
\label{sec:appendix_clicksim}

Annotating real interactive sessions at scale is impractical, so we simulate the ``user'' from the ground-truth mask, exactly as in SAM-style interactive evaluation~\cite{kirillov2023segment,zhou2025pointsam}.
The \emph{same} rollout (Algorithm~\ref{alg:clicksim}) is used for training and evaluation, so the model is optimized under the protocol it is scored with.
For a target tree with mask $\mathbf{m}^{\star}$ over the scene points (voxelized for decoding), the rollout proceeds for $T$ iterations.

\vspace{4pt}
\noindent\textbf{First click.}
Because the headline comparison is decided at a single click, we specify its placement exactly.
Let $\mathcal{I}=\{i:\mathbf{m}^{\star}_i{=}1\}$ index the points of the target tree and $\bar{\mathbf{p}}$ their centroid.
The first click is the in-mask point closest to that centroid:
\begin{equation}
\label{eq:first_click}
\mathbf{c}_{\mathrm{in}}=\mathbf{p}_{i^{\star}},~~
\bar{\mathbf{p}}=\frac{1}{|\mathcal{I}|}\sum_{i\in\mathcal{I}}\mathbf{p}_i,~~
i^{\star}=\operatorname*{arg\,min}_{i\in\mathcal{I}}\ \|\mathbf{p}_i-\bar{\mathbf{p}}\|_2 .
\end{equation}
Three properties follow.
It is \emph{deterministic}: no sampling, no seed, and no jitter is applied, so a run is reproducible and the reported numbers carry no click-sampling variance.
It is \emph{shared}: every compared method receives this identical click for the same tree, so the protocol cannot favour one model over another.
And it is \emph{favourable}: the centroid-nearest point is the most interior point of the mask, which for a tree typically lies on the stem or in the inner crown rather than near an entangled crown boundary.
This follows the center-based initial click adopted by prior interactive 3D segmentation work, and is what makes single-click numbers comparable across papers, but it is a best-case placement rather than a model of where a human would click.
Absolute IoU@1 should therefore be read as an upper bound for all methods alike; the effect of less favourable placements, and of real human clicks, is discussed in the limitations of the main paper.

\vspace{4pt}
\noindent\textbf{CHM treetop click.}
This is the only place the ground-truth mask enters the first prompt.
In the \mbox{+ w/ CHM} setting it is supplemented by a treetop click $\mathbf{c}_{\mathrm{top}}$.
The candidate treetops are detected from the CHM over the whole scene before any click is placed, and the one prompted here is selected from that scene-level set by the geometric association rule of the main paper, which prompts the nearest peak whose allometric crown footprint covers $\mathbf{c}_{\mathrm{in}}$ and uses only that click coordinate.
The mask is used neither to detect the peaks nor to choose among them, so the treetop click can land on a neighbouring crown; when no candidate covers the click, the first prompt stays the interior click alone.

\vspace{4pt}
\noindent\textbf{Correction clicks.}
At each subsequent iteration we compute the error between the current prediction $\hat{\mathbf{m}}^{(t)}$ and $\mathbf{m}^{\star}$, splitting it into a false-negative (missed) region $\mathcal{F}^{-}{=}\mathbf{m}^{\star}\wedge\neg\hat{\mathbf{m}}^{(t)}$ and a false-positive (leaked) region $\mathcal{F}^{+}{=}\neg\mathbf{m}^{\star}\wedge\hat{\mathbf{m}}^{(t)}$.
We correct the larger error first: if $|\mathcal{F}^{-}|\!\ge\!|\mathcal{F}^{+}|$ we add a \emph{positive} click at the most interior point of $\mathcal{F}^{-}$, otherwise a \emph{negative} click at the most interior point of $\mathcal{F}^{+}$.
Selecting the most interior point (farthest from the region boundary) yields a stable, deterministic click that mirrors how a user corrects an obvious mistake.
Clicks accumulate across iterations and the query is rebuilt from the full click set each step.

\vspace{4pt}
\noindent\textbf{Counting policy.}
Throughout the paper the click budget counts only the simulated user clicks, namely the interior click and the correction clicks.
The CHM treetop is provided in addition and does not count toward the budget, so at every budget all methods receive the same number of user clicks.
During training we run $T{=}5$ iterations per tree and supervise the decoded mask at each iteration; at evaluation the rollout produces the IoU@$k$ and NoC curves reported in the main paper.
Evaluation runs the rollout to a horizon of $10$ clicks; a tree that never reaches an IoU target within that horizon is counted at $10$ clicks for that target, so NoC is bounded and comparable across methods.

\section{Dataset Details}
\label{sec:appendix_dataset}

\noindent\textbf{FOR-instanceV2.}
Following ForestFormer3D~\cite{xiang2025forestformer3d} and ForestMamba~\cite{nguyen_forestmamba}, our in-distribution experiments use the FOR-instanceV2 benchmark~\cite{xiang2025forestformer3d}, a collection of close-range forest LiDAR point clouds from seven geographically diverse regions across four continents, summarized in Table~\ref{tab:appendix_dataset}.
The regions span boreal, temperate, and tropical forest types and are acquired with Unmanned (ULS), Terrestrial (TLS), and Mobile (MLS) Laser Scanning systems, giving substantial variation in forest structure, point density, terrain, and acquisition geometry.
Every point carries both a semantic label (ground, wood, leaf) and an individual tree-instance ID.
In total the dataset provides $8{,}016$ annotated tree instances for training, $1{,}374$ for validation, and $1{,}744$ for testing ($11{,}134$ trees overall).
Only the raw $(x,y,z)$ coordinates are used as input, without color or intensity.

\vspace{4pt}
\noindent\textbf{LAUTx (cross-dataset generalization).}
To probe generalization beyond the training distribution, we additionally evaluate on LAUTx~\cite{tockner2022lautx}, a set of $516$ individual-tree point clouds collected from Austrian forest inventory plots.
LAUTx is \emph{never} seen during training or validation and differs from FOR-instanceV2 in geographic location, plot composition, and acquisition characteristics, making it a strict test of cross-dataset transfer.
We apply the same promptable evaluation protocol (Sec.~\ref{sec:appendix_clicksim}) and the same input pre-processing as for FOR-instanceV2, without any re-tuning or fine-tuning on LAUTx.

\section{Implementation Details}
\label{sec:appendix_impl}

\noindent\textbf{Framework.}
SelectAnyTree is implemented on top of the MMDetection3D framework, with the promptable path, prompt encoder, click simulation, and interactive metrics added as custom modules.
{The backbone and state-space query decoder are inspired by the structure-aware forest model~\cite{nguyen_forestmamba}, and are specialized for the promptable path by the click-simulation training.}

\vspace{4pt}
\noindent\textbf{Data preprocessing.}
Each plot is cropped into cylindrical patches of radius $16$\,m, grid-sampled at $0.2$\,m, and subsampled to at most $640$k points, the same resolution used for voxelization.
We use only the raw $(x,y,z)$ coordinates as input, without color or intensity, so the model relies purely on geometry.
At validation and test time each plot is processed without cropping, and the scene is encoded once and reused across all prompts, which is what makes interactive multi-tree selection efficient.

\vspace{4pt}
\noindent\textbf{Data augmentation.}
During training we apply random horizontal flipping, random rotation about the vertical axis, and uniform scaling in $[0.8, 1.2]$, together with light coordinate jitter.
These augmentations preserve the ground-to-canopy vertical structure that the backbone and the CHM prompt rely on, while improving robustness to plot orientation and scale.

\vspace{4pt}
\noindent\textbf{Training and inference.}
We train with a batch size of one plot using synchronized batch normalization across GPUs, validating every $50$ epochs with the same click-simulation rollout used for the final evaluation.
The promptable mask loss is supervised at every rollout iteration, so a single forward pass trains all click counts jointly.
All experiments, including training and the timing measurements reported in the main paper, were run on an NVIDIA A6000 GPU.

{\section{Baseline Implementation and Fine-tuning}}
\label{sec:appendix_baselines}

\begin{table*}[th]
\centering
\caption{Hyperparameter summary for the proposed SelectAnyTree.}
\label{tab:appendix_hparams}
{
\begin{tabular}{lll}
\toprule
Component & Parameter & Value \\
\midrule
\multirow{4}{*}{Encoder}
  & Voxel size $v$              & 0.2\,m \\
  & Backbone width $C$          & 32 \\
  & Mamba $d_{\mathrm{state}}$  & 16 \\
  & Slab thickness $\tau$       & 5 voxels \\
\midrule
\multirow{5}{*}{Decoder}
  & Layers $L$                  & 6 \\
  & Model dim.\ $D'$            & 256 \\
  & Decoder $d_{\mathrm{state}}$ & 64 \\
  & FFN hidden dim.             & 1,024 \\
  & Mask-feedback rank $r$ & 16 \\
\midrule
\multirow{2}{*}{Prompt encoder}
  & Fourier half-width $d_{\mathrm{pe}}$ & 64 \\
  & Geometric anchor   & nearest voxel \\
\midrule
\multirow{4}{*}{CHM prompt}
  & CHM scales                  & \{0.3\,m, 0.7\,m\} \\
  & Allometric $(\alpha,\beta)$ & (0.25, 0.5) \\
  & Min.\ height / separation   & 1.5\,m / 1.0\,m \\
  & Association   & nearest covering peak \\
\midrule
\multirow{3}{*}{Click simulation}
  & Iterations $T$ (train)      & 5 \\
  & First prompt                & interior click ($+$ CHM treetop optional) \\
  & Correction                  & error-region (pos./neg.) \\
\midrule
\multirow{5}{*}{Training}
  & Epochs                      & 1,000 \\
  & Optimizer                   & AdamW~\cite{loshchilov2019decoupled} \\
  & Learning rate / decay       & $10^{-4}$ / poly (0.9) \\
  & Weight decay / grad.\ clip  & 0.05 / 10 \\
  & Max prompts / scene         & 128 \\
\midrule
\multirow{2}{*}{Loss weights}
  & $(\lambda_{\mathrm{p}},\lambda_{\mathrm{bin}},\lambda_{\mathrm{dis}})$ & (1.0, 1.0, 1.0) \\
  & Dice weight $\lambda_{\mathrm{dice}}$ & 2.0 \\
\bottomrule
\end{tabular}
}
\end{table*}

{\noindent\textbf{Overview and fairness protocol.}
We compare against four promptable 3D segmenters: AGILE3D~\cite{yue2024agile3d}, NPISeg3D~\cite{liu2025npising3d}, Point-SAM~\cite{zhou2025pointsam}, and PartSAM~\cite{zhu2026partsam}.
None of these models was trained on forest LiDAR, so for a fair in-distribution comparison every baseline is trained on the FOR-instanceV2~\cite{xiang2025forestformer3d} benchmark, from scratch where no suitable checkpoint exists and by fine-tuning the released weights otherwise, as detailed below.
Each baseline is trained and evaluated on the \emph{same} scenes and the \emph{same} per-tree ground-truth instance masks as SelectAnyTree, and scored with an identical metric implementation under a GT-driven, human-free click-simulation protocol.}
Every method therefore sees the same input: the same plots, the same trees, the same clicks, and the same evaluation.
What differs is only what each architecture does with that input internally.
AGILE3D and NPISeg3D consume the plot at a $0.4$\,m voxel, Point-SAM and PartSAM subsample it to the point count of their released pipelines, and SelectAnyTree voxelizes at $0.2$\,m; in each case the setting is the one the method was designed and released with, not a budget we assigned.
These internal resolutions are not comparable as a single number, so we do not present one as an advantage.
Predicted masks are mapped back to the original points before scoring, so IoU is computed over the identical point set for every method and no model is credited or penalised for the resolution at which it works.

\vspace{4pt}
{\noindent\textbf{AGILE3D~\cite{yue2024agile3d}.}
AGILE3D is a MinkowskiEngine-based multi-object interactive segmenter built for small indoor scans.
We train it from scratch on FOR-instanceV2 using its native multi-object regime: each forest plot is taken as a whole scene (no cylinder cropping), subsampled to $400$k points, and voxelized at $0.4$\,m.
The coarser $0.4$\,m voxel (vs.\ the $0.05$\,m used indoors) is required because forest plots are meter-scale and because AGILE3D's click simulator computes a background-to-foreground distance matrix whose memory grows quadratically with voxel count; we additionally chunk this distance computation over foreground points, which is mathematically identical but keeps memory bounded on large plots.
Up to ten trees per scene are segmented jointly, matching its original multi-object setting.
Clicks are simulated from the ground truth at both training and evaluation time, and we train with AdamW, learning rate $10^{-4}$, batch size $4$, for $1{,}000$ epochs with validation every $50$ epochs.}

\vspace{4pt}{\noindent\textbf{NPISeg3D~\cite{liu2025npising3d}.}
NPISeg3D extends AGILE3D with a hierarchical neural process that produces probabilistic masks and per-click uncertainty.
We reuse exactly the same converted FOR-instanceV2 data, voxel size ($0.4$\,m), and click-simulation protocol as the AGILE3D baseline, and train it under the same schedule (AdamW, learning rate $10^{-4}$, batch size $5$, $\sim$$1{,}100$ epochs, checkpointing every $50$ epochs).
Because NPISeg3D's recurrent mask encoder is fixed to a background slot plus a small number of foreground objects, we cap the number of trees per crop accordingly; at inference the predicted mask is obtained by averaging over the neural-process samples before the per-voxel argmax.}

\vspace{4pt}
{\noindent\textbf{Point-SAM~\cite{zhou2025pointsam}.}
Point-SAM is the general-purpose promptable 3D segmenter and uses a ViT-L point-cloud encoder with a SAM-style prompt encoder and mask decoder.
We fine-tune it from the released pretrained checkpoint on FOR-instanceV2.
Inputs follow Point-SAM's own pipeline: coordinates are normalized to $[-1,1]$ and the colour channels are set to zero, since forest LiDAR carries no RGB and the model relies purely on geometry.
We feed cylindrical crops subsampled to $\sim$$20$k points, simulate clicks from the ground truth using Point-SAM's native scheme (an in-mask first click followed by farthest-from-border error-region corrections), and optimise the SAM-style mask loss (BCE + Dice) with AdamW at learning rate $10^{-4}$, batch size $1$, for $1{,}000$ epochs with validation every $50$ epochs.
}

\vspace{4pt}
{\noindent\textbf{PartSAM~\cite{zhu2026partsam}.}
PartSAM is built directly on the Point-SAM code base and adds a PartField triplane feature encoder for part-level geometry.
Its released code exposes only an inference path; the multi-round click-simulation training loop is not published, so we replicate Point-SAM's interactive training loop driving PartSAM's \texttt{predict\_masks}, and supervise it with PartSAM's own criterion.
The triplet/contrastive term is omitted because the three-view augmentation pipeline it depends on is unreleased.
We fine-tune from the released pretrained weights on the same cylindrical FOR-instanceV2 crops as Point-SAM, with coordinates normalized to $[-1,1]$ and both colour and surface normals set to zero (the PartField geometric features still flow from the coordinates); training uses AdamW at learning rate $10^{-4}$, batch size $1$, for $1{,}000$ epochs with validation every $50$ epochs.}

\vspace{10pt}
\section{Hyperparameters}
\label{sec:appendix_hparams}

Table~\ref{tab:appendix_hparams} lists the hyperparameters of SelectAnyTree.
A single configuration is shared across all seven forest regions and reused unchanged on the cross-dataset LAUTx evaluation, without any per-region or per-dataset tuning.
The backbone and decoder dimensions follow the structure-aware forest backbone~\cite{nguyen_forestmamba}, while the prompt encoder, CHM prompt, and click-simulation settings are specific to the promptable path introduced in this work.

\vspace{4pt}
\noindent\textbf{Backbone and decoder.}
The point cloud is voxelized at $v{=}0.2$\,m, which balances geometric detail against the number of occupied voxels.
The five-level sparse encoder has a base width of $32$ channels and a Mamba state dimension of $16$, with a slab thickness of $\tau{=}5$ voxels (roughly $1$\,m in height) so that each serialized slab groups vertically coherent structure.
The state-space query decoder uses $L{=}6$ layers, a model dimension of $256$, a decoder state dimension of $64$, and a feed-forward hidden dimension of $1,024$, matching common practice in query-based 3D segmentation~\cite{kolodiazhnyi2024oneformer3d,xiang2025forestformer3d}.
The mask feedback of the main paper adds a low-rank projection $\mathbf{W}_h\in\mathbb{R}^{r\times C}$ with $r{=}16$ and a gate MLP $\psi$ on top of the decoder.
The gate is zero-initialized, so training starts from the feedback-free decoder and the feedback path is learned rather than imposed.

\begin{table*}[t]
\centering
\caption{{Bootstrap mean\,$\pm$\,std of the interactive metrics on the in-distribution FOR-instanceV2~\cite{xiang2025forestformer3d} test set ($B{=}2000$ resamples over test-set trees, single checkpoint). IoU is in \%; NoC is the number of user clicks (lower is better).}}
\label{tab:appendix_bootstrap}
{%
\begin{tabular}{l|ccc|ccc}
\toprule
\multirow{2}{*}{Method} & \multicolumn{3}{c|}{IoU @ Clicks $\uparrow$} & \multicolumn{3}{c}{NoC @ IoU $\downarrow$} \\
\cmidrule(lr){2-4} \cmidrule(lr){5-7}
 & 1 & 3 & 5 & 0.7 & 0.8 & 0.9 \\
\midrule
SelectAnyTree (Proposed)
 & 77.9\,$\pm$\,0.62 & 81.3\,$\pm$\,0.60 & 81.1\,$\pm$\,0.59
 & 3.0\,$\pm$\,0.10 & 3.9\,$\pm$\,0.12 & 5.5\,$\pm$\,0.13 \\
\hspace{5pt}+ w/ CHM
 & 79.9\,$\pm$\,0.64 & 81.9\,$\pm$\,0.61 & 82.4\,$\pm$\,0.60
 & 2.9\,$\pm$\,0.10 & 3.8\,$\pm$\,0.12 & 5.4\,$\pm$\,0.13 \\
\bottomrule
\end{tabular}%
}
\end{table*}

\vspace{4pt}
\noindent\textbf{Prompt encoder and CHM prompt.}
The prompt encoder uses a random-Fourier positional encoding of half-width $d_{\mathrm{pe}}{=}64$ and reads the geometric anchor from the backbone feature of the voxel nearest each positive click.
Averaging over a cylinder instead blurs the anchor across adjacent crowns, as the ablation in the main paper shows.
The CHM prompt detects treetops at two grid resolutions ($0.3$\,m and $0.7$\,m) to cover both small suppressed and large dominant crowns, with a minimum tree height of $1.5$\,m to reject ground clutter and a minimum peak separation of $1.0$\,m to keep closely spaced trees distinct.
The allometric parameters $(\alpha,\beta){=}(0.25,0.5)$ follow the ecologically grounded power-law relationship between crown radius and tree height~\cite{jucker2017allometric}, and define the crown radius $r_k{=}\alpha h_k^{\beta}$ that decides which detected peaks are eligible for a given click.
Among the eligible peaks, the one nearest the click in the horizontal plane is prompted; if none is eligible, no treetop click is issued.

\vspace{4pt}
\noindent\textbf{Click simulation and training.}
We run $T{=}5$ rollout iterations per tree during training, with the first prompt comprising the interior click, optionally supplemented by the CHM treetop in the \mbox{+ w/ CHM} setting, and each subsequent click sampled from the prediction error region.
At most $128$ trees are prompted per scene to bound memory, and the decoded mask is supervised at every iteration.
We train for $1{,}000$ epochs with AdamW~\cite{loshchilov2019decoupled} ($\mathrm{lr}{=}10^{-4}$, weight decay $0.05$) under a polynomial schedule (power $0.9$) with gradient clipping at norm $10$.
The mask, binary, and discriminative losses are weighted equally ($\lambda_{\mathrm{p}}{=}\lambda_{\mathrm{bin}}{=}\lambda_{\mathrm{dis}}{=}1.0$), and within the mask loss the Dice term is weighted at $\lambda_{\mathrm{dice}}{=}2.0$ to counteract the strong foreground/background imbalance of segmenting one tree at a time.

\section{Statistical Robustness}
\label{sec:appendix_bootstrap}
{The click-simulation rollout is deterministic, so repeating the evaluation of a trained model yields identical numbers and zero run-to-run variance.
To quantify the uncertainty of the reported metrics from a single checkpoint, we instead bootstrap over the test-set trees: the per-tree IoU curves are resampled with replacement $B{=}2000$ times, and for each resample we recompute the interactive metrics.
Table~\ref{tab:appendix_bootstrap} reports the resulting mean and standard deviation on the in-distribution FOR-instanceV2 test set, for SelectAnyTree with and without the CHM prompt.
The standard deviations are small ($\le 0.64$ points for IoU and $\le 0.13$ clicks for NoC), confirming that the few-click advantage of SelectAnyTree is stable across the test population rather than driven by a handful of easy trees.}




\section{Additional Qualitative Results}
\label{sec:appendix_qual}

\begin{figure*}[t]
  \centering
  \begin{subfigure}{1.0\linewidth}
    \centering
    \includegraphics[width=\textwidth]{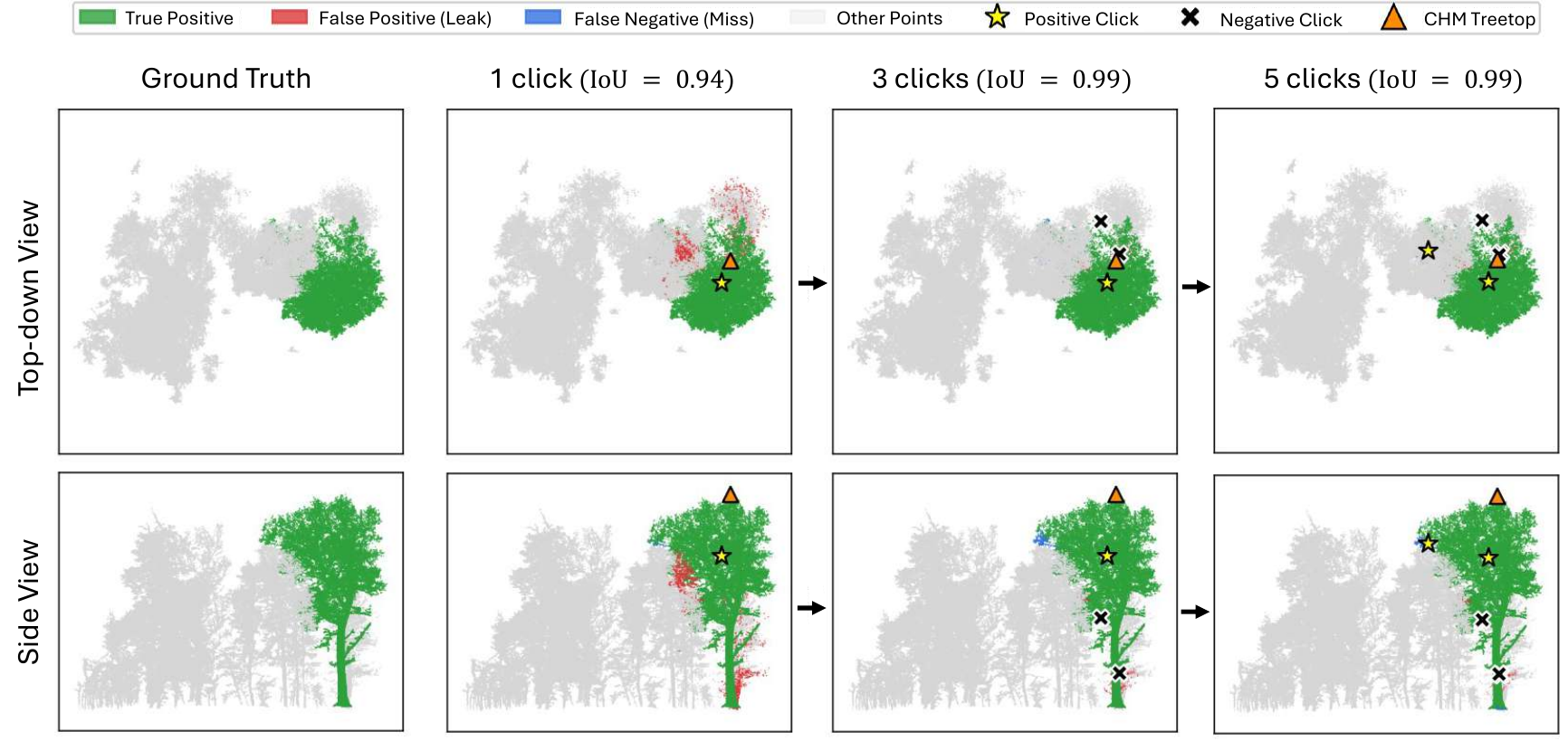}
    \subcaption{BlueCat (Czech Republic)~\cite{xiang2025forestformer3d}: A dense temperate forest with strongly overlapping crowns.}
    \label{fig:appendix_bluecat}
  \end{subfigure}
  \begin{subfigure}{1.0\linewidth}
    \centering
    \includegraphics[width=\textwidth]{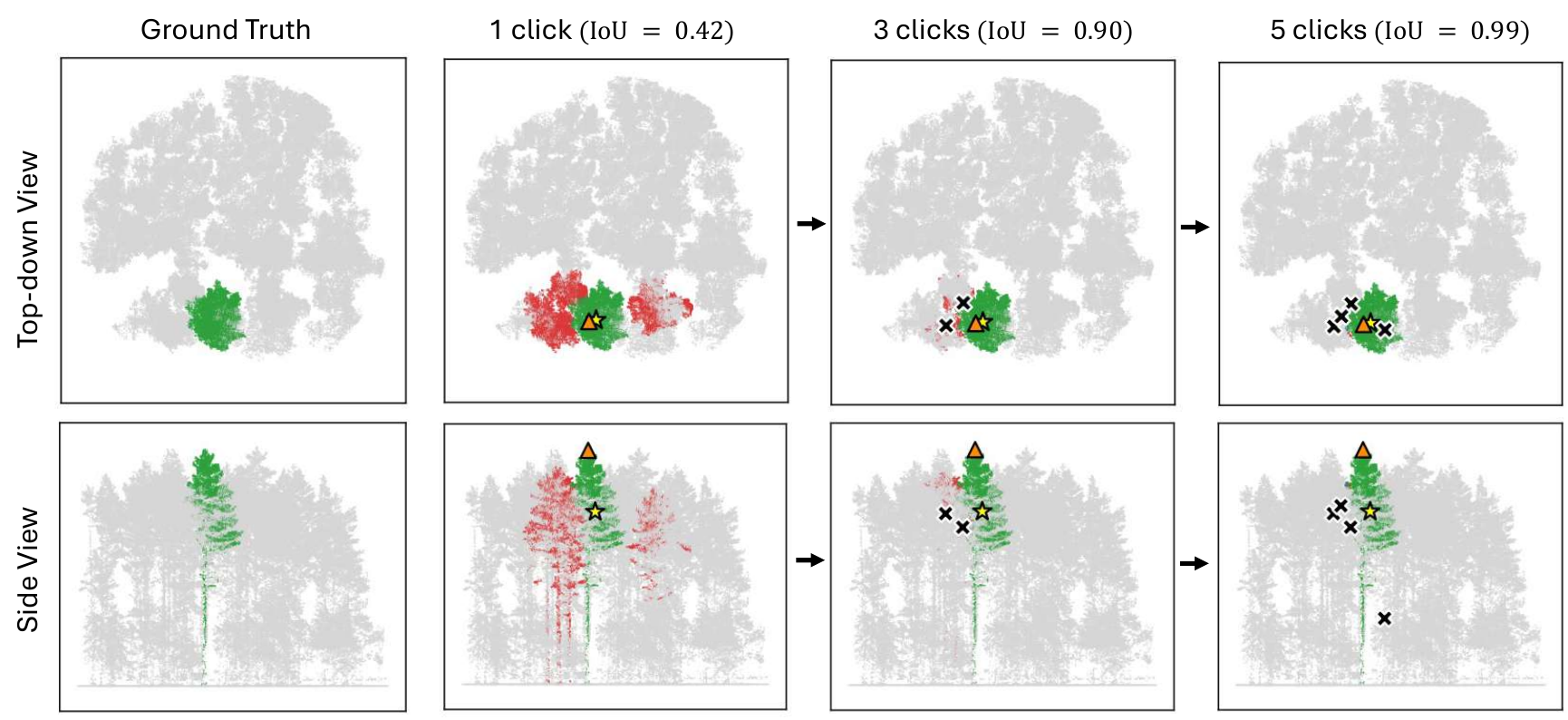}
    \subcaption{NIBIO (Norway)~\cite{xiang2025forestformer3d}: A boreal conifer plot.}
    \label{fig:appendix_nibio}
  \end{subfigure}
  \caption{Promptable tree segmentation on FOR-instanceV2~\cite{xiang2025forestformer3d}.
  Given a few clicks on a target tree, SelectAnyTree segments that individual tree within dense, overlapping canopies.
  For each example we show the input clicks, the predicted mask, and the ground truth.
  Positive and negative clicks are marked, color is assigned per instance, and ground points are omitted for clarity.}
  \label{fig:appendix_qual}
\end{figure*}

\begin{figure*}[t]
  \centering
  \begin{subfigure}{1.0\linewidth}
    \centering
    \includegraphics[width=\textwidth]{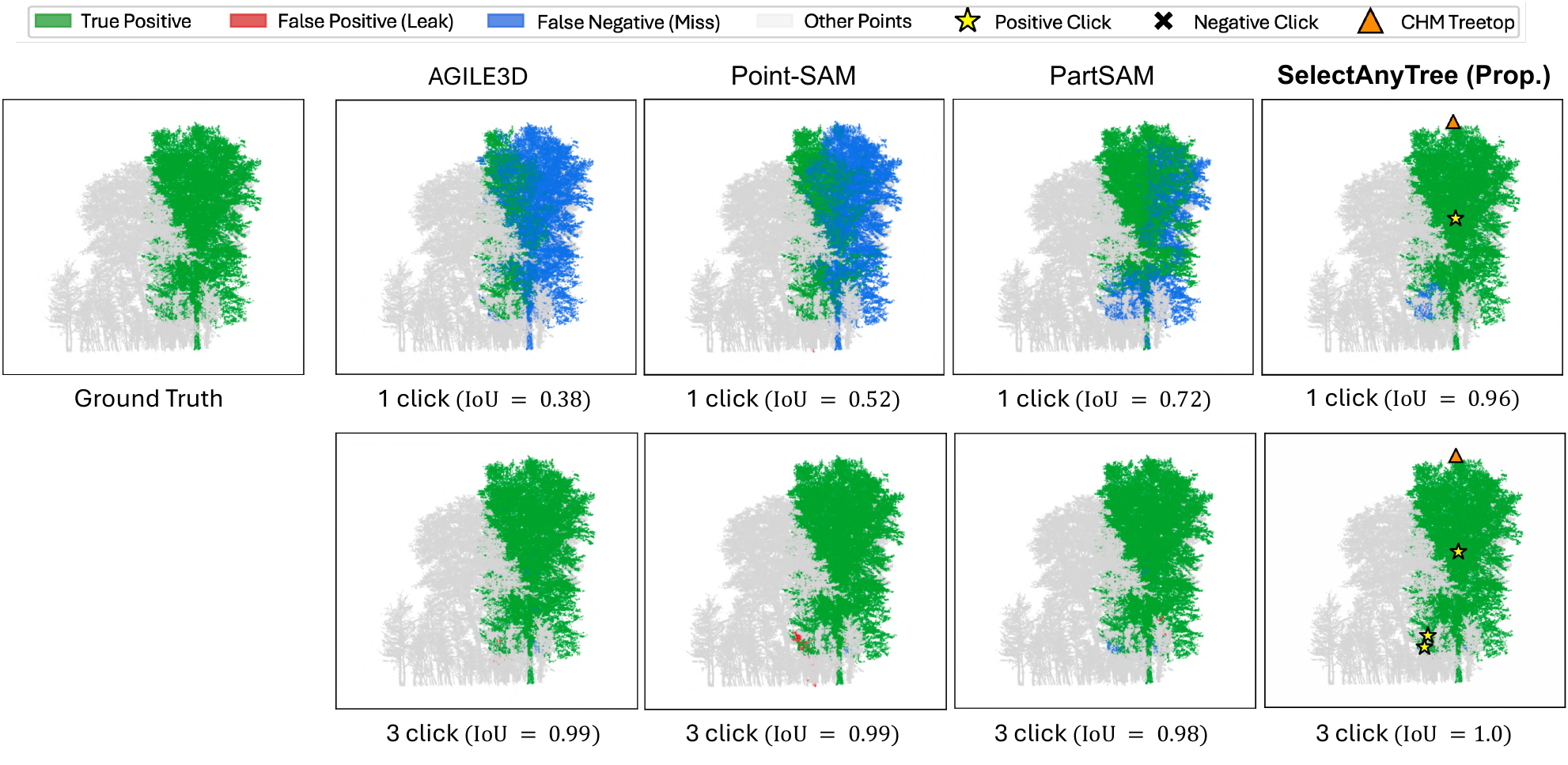}
    \subcaption{BlueCat (Czech Republic): A dense temperate forest with strongly overlapping crowns.}
    \label{fig:qualitative_bluecat_appendix}
  \end{subfigure}
  \begin{subfigure}{1.0\linewidth}
    \centering
    \includegraphics[width=\textwidth]{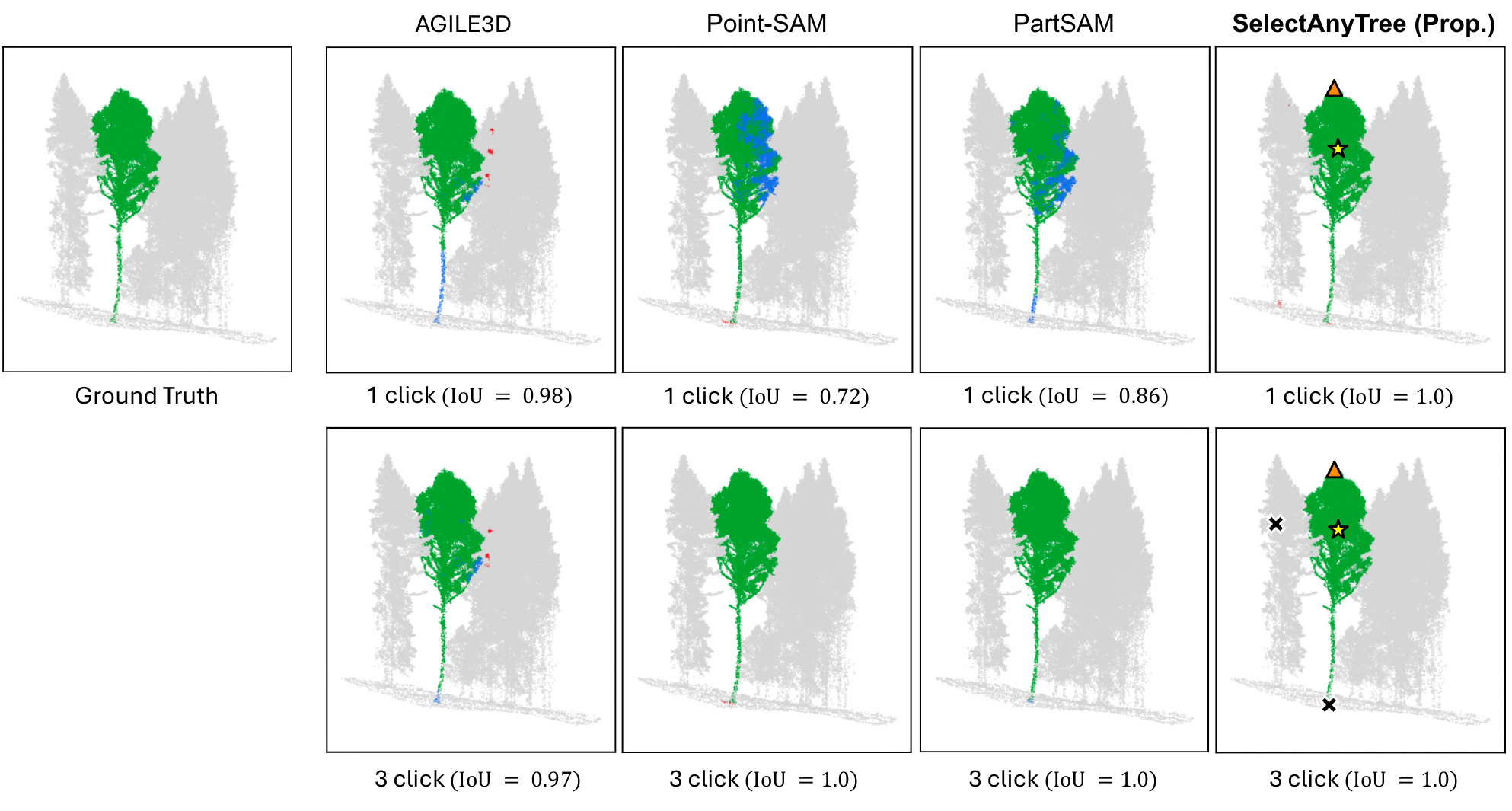}
    \subcaption{NIBIO (Norway): A boreal conifer plot.}
    \label{fig:qualitative_nibio_appendix}
  \end{subfigure}
  \caption{Single-click tree segmentation.
    Comparison of the three strongest promptable baselines,
    AGILE3D~\cite{yue2024agile3d}, Point-SAM~\cite{zhou2025pointsam}, and
    PartSAM~\cite{zhu2026partsam}, against the proposed SelectAnyTree.}
  \label{fig:qualitative_comparison_appendix}
\end{figure*}

Figure~\ref{fig:appendix_qual} provides qualitative examples on the dense BlueCat (Czech Republic) and boreal NIBIO (Norway) regions, complementing the single-click comparison in the main paper.
For each target tree we show both a top-down and a side view at increasing click budgets, so that horizontal separation from neighbouring crowns and vertical recovery of the full stem-to-canopy extent can be inspected together.
Points are colored as true positives (green), false positives or leaks (red), and false negatives or misses (blue), and the simulated clicks are overlaid as a positive click (star), a negative click (cross), and the CHM treetop (triangle).

On the dense temperate BlueCat plot (Fig.~\ref{fig:appendix_bluecat}), the first prompt alone, the interior click together with the CHM treetop, already segments the target tree almost completely at $0.94$ IoU, leaving only a small leak into a strongly overlapping neighbour visible as a red patch in the top-down view.
A single negative click placed on that leaked region removes it, and the mask sharpens to $0.99$ IoU at three clicks and remains stable at five, with the side view confirming that the entire trunk and crown are recovered without spilling onto adjacent stems.

The boreal NIBIO plot (Fig.~\ref{fig:appendix_nibio}) is more challenging.
Here the first prompt leaks substantially into two neighbouring conifers whose crowns interlock with the target, yielding a low $0.42$ IoU dominated by red false positives that the side view reveals to run along the adjacent vertical stems.
As the simulated user adds negative clicks precisely on these leaked stems, the prediction is progressively carved back to the true tree, climbing to $0.90$ IoU at three clicks and $0.99$ at five.
Across both regions the CHM treetop consistently anchors the canopy apex from the first iteration, the interior click grounds the trunk, and subsequent negative clicks act locally to suppress leakage rather than perturbing the whole mask, which is exactly the behaviour that makes the model efficient to correct.
These examples reinforce the trend shown in the figure in the main paper, where SelectAnyTree starts from a much stronger single-click mask than AGILE3D, Point-SAM, and PartSAM, and reaches near-perfect segmentation within only a few interactions.